\documentclass[10pt,journal,compsoc]{IEEEtran}

\ifCLASSOPTIONcompsoc
  \usepackage[nocompress]{cite}
\else
  \usepackage{cite}
\fi

\usepackage{graphicx}
\graphicspath{{.}{./Image/}}
\usepackage{subfigure}

\usepackage{amsmath,amssymb,amsfonts}
\usepackage{xspace}
\usepackage{booktabs}
\usepackage[ruled,vlined,linesnumbered]{algorithm2e}
\usepackage{url}
\usepackage{wrapfig}
\usepackage{pifont}
\newcommand{\cmark}{\text{\ding{51}}}
\newcommand{\xmark}{\text{\ding{55}}}

\newcommand{\shortname}{FedTune\xspace}

\begin{document}

\title{Federated Learning Hyper-Parameter Tuning From A System Perspective}

\author{Huanle~Zhang,
       Lei~Fu,
       Mi~Zhang,
       Pengfei~Hu,
       Xiuzhen~Cheng,~\IEEEmembership{Fellow,~IEEE},
       Prasant~Mohapatra,~\IEEEmembership{Fellow,~IEEE},
       and~Xin~Liu,~\IEEEmembership{Fellow,~IEEE},
\IEEEcompsocitemizethanks{\IEEEcompsocthanksitem 
Huanle Zhang, Pengfei Hu, and Xiuzhen Cheng are with the School of Computer Science and Technology, Shandong University, China.\protect\\
E-mail: \{dtczhang, phu, xzcheng\}@sdu.edu.cn 
\IEEEcompsocthanksitem 
Lei Fu is with the Bank of Jiangsu and Fudan University, China. \protect\\
E-mail: leileifu@163.sufe.edu 
\IEEEcompsocthanksitem 
Mi Zhang is with the Department of Computer Science and Engineering, Ohio State University, USA. \protect\\
E-mail: mizhang.1@osu.edu
\IEEEcompsocthanksitem 
Prasant Mohapatra and Xin Liu are with the Department of Computer Science, University of California, Davis, USA. \protect\\
E-mail: \{pmohapatra,xinliu\}@ucdavis.edu
}
\thanks{Pengfei Hu is the corresponding author.}
}

%
%

\markboth{IEEE Transactions on Mobile Computing}%
{Shell \MakeLowercase{\textit{et al.}}: Bare Demo of IEEEtran.cls for Computer Society Journals}
%



\IEEEtitleabstractindextext{%
\begin{abstract}
Federated learning (FL) is a distributed model training paradigm that preserves clients' data privacy. It has gained tremendous attention from both academia and industry. 
FL hyper-parameters (e.g., the number of selected clients and the number of training passes) significantly affect the training overhead in terms of computation time, transmission time, computation load, and transmission load.
However, the current practice of manually selecting FL hyper-parameters imposes a heavy burden on FL practitioners because applications have different training preferences. In this paper, we propose \shortname, an automatic FL hyper-parameter tuning algorithm tailored to applications' diverse system requirements in FL training.
\shortname iteratively adjusts FL hyper-parameters during FL training and can be easily integrated into existing FL systems. Through extensive evaluations of \shortname for diverse applications and FL aggregation algorithms,  we show that \shortname is lightweight and effective, achieving 8.48\%-26.75\% system overhead reduction  compared to using fixed FL hyper-parameters. This paper assists FL practitioners in designing high-performance FL training solutions. 
The source code of \shortname is available at \url{https://github.com/DataSysTech/FedTune}.
\end{abstract}

}

\maketitle

\IEEEdisplaynontitleabstractindextext

\IEEEpeerreviewmaketitle

\IEEEraisesectionheading{\section{Introduction}\label{sec:introduction}}

\IEEEPARstart{F}{ederated} Learning (FL) has been applied to a wide range of applications such as mobile keyboard~\cite{gboard}, speech recognition~\cite{appleSpeech}, and human stroke prevention~\cite{stroke} on top of mobile devices and Internet of Things (IoT)~\cite{dledge2020bookchapter,fliotvision2022ieeeiotm}.
Compared to other model training paradigms~(e.g., centralized machine learning~\cite{machine_learning}, conventional distributed machine learning~\cite{distributed_learning}), FL has unique properties with regards to
(1) \textit{Massively Distributed}: the number of clients is much larger than the clients' average number of data points; (2) \textit{Unbalanced Data}: clients have a different amount of data points; and (3) \textit{Non-IID Data}: each client's data may not represent the overall distribution~\cite{FedAvg17aistats}. 
In addition to the common hyper-parameters of model training such as learning rates, optimizers, and mini-batch sizes, FL has unique hyper-parameters, including aggregation methods and participant selection~\cite{Oort21OSDI,pyramidfl2022mobicom}. Nonetheless, many FL algorithms, e.g., FedAvg~\cite{FedAvg17aistats}, have been proved to converge to the global optimum even different FL hyper-parameters are adopted~\cite{fedAvgConverge20iclr}\cite{fedNova20neurips}.

\begin{figure}[!t]
    \centering
    \includegraphics[width=3.2in]{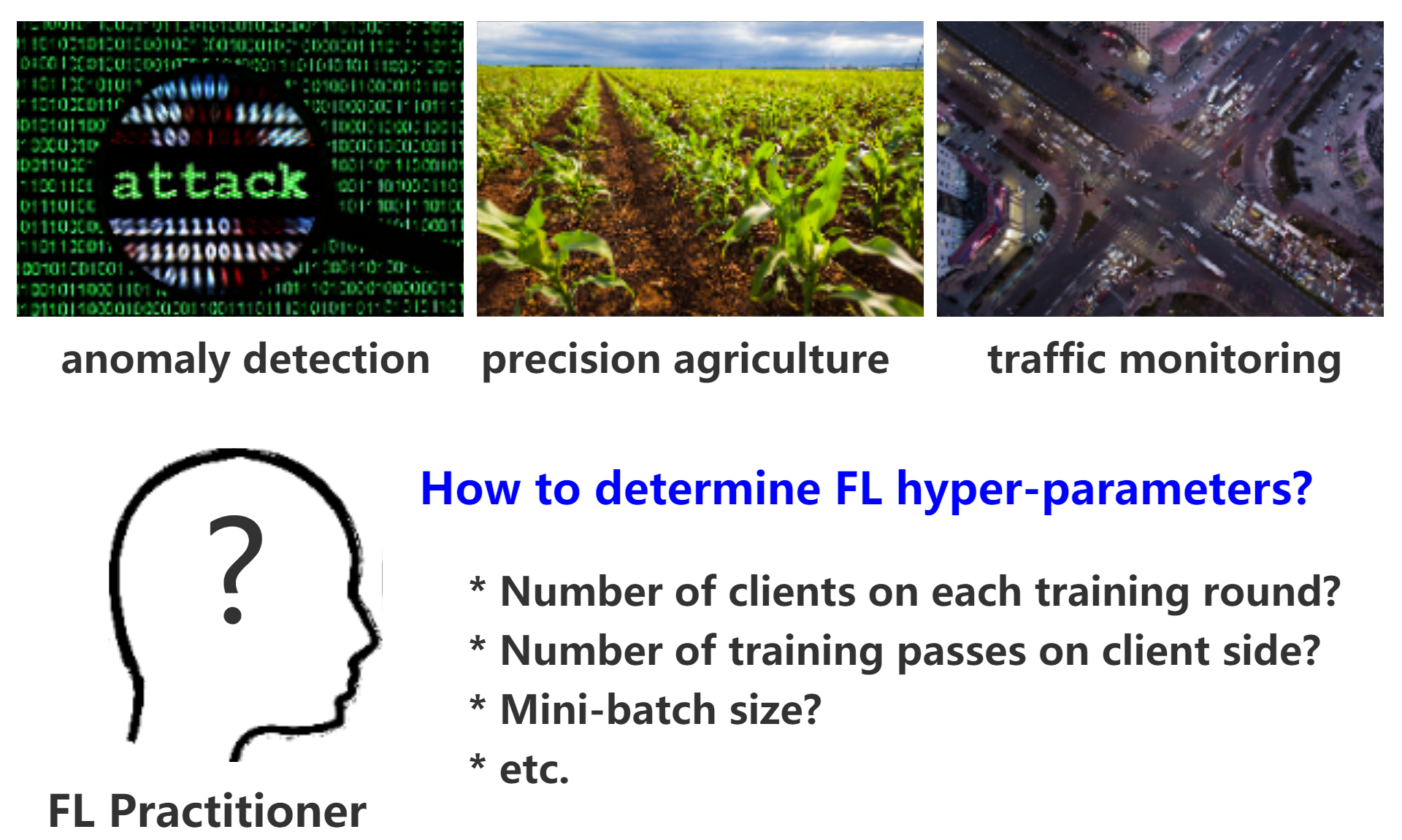}
    \caption{FL practitioners have various constraints in determining FL hyper-parameters for different applications.}
    \label{fig:fl_practitioner}
\end{figure}

Although FL hyper-parameters do not affect FL convergence (i.e., the same final global model accuracy), they significantly affect the training overheads of reaching the final model. Specifically, Computation Time (CompT), Transmission Time (TransT), Computation Load (CompL), and Transmission Load (TransL) are the four most crucial system overhead:
\begin{itemize}
    \item \textit{Computation Time (CompT)}. It measures how long an FL system spends in model training.  
    Overall model training time must be short for applications that require fast adaptation to environments (e.g., security problems).  
    \item \textit{Transmission Time (TransT)}. 
    It represents how long an FL system spends in model parameter transmission between clients and servers. Without a high-speed transmission solution, transmission time can overwhelm the overall FL training since models are generally large and multiple rounds of model parameter transmissions are required. 
    
    \item \textit{Computation Load (CompL)}. It is
    the number of Floating-Point Operation (FLOP) that an FL system consumes.
    For low-profile devices such as IoT nodes, the computation load must be small as these devices are equipped with low computational resources.

    \item \textit{Transmission Load (TransL)}. It is the total data size transmitted between the clients and the server. This is critical when data transmission is expensive or its power consumption is a concern. For example, outdoor applications usually rely on cellular communications, which need to pay a considerable price for transmitting a large amount of data. 
\end{itemize}

Fig.~\ref{fig:fl_practitioner} summarizes the different application scenarios faced by FL practitioners in determining FL hyper-parameters. Application scenarios have different training preferences in terms of CompT, TransT, CompL, and TransL. Consider the following examples: (1) attack and anomaly detection in computer networks~\cite{anomaly21ajrcos}
is time-sensitive (CompT and TransT) as it needs to adapt to malicious traffic rapidly; (2) smart home control systems for indoor environment automation~\cite{smart_home21jaihc}, e.g., heating, ventilation, and air conditioning (HVAC), are sensitive to computation (CompT and CompL) because sensor devices are limited in computation capabilities; (3) traffic monitoring systems for vehicles~\cite{traffic20access}
are communication-sensitive (TransT and TransL) because cellular communications are usually adopted to provide city-scale connectivity; 
(4) precision agriculture based on IoT sensing~\cite{agriculture20access}
is not time-urgent but requires energy-efficient solutions (CompL and TransL); (5) healthcare systems, e.g., fall detection for elderly people~\cite{fall19access}, require both fast response and small energy consumption (CompT, TransT, CompL, and TransL); and (6) human stampedes detection/prevention~\cite{stampede18pdm} require time, computation, and communication efficient systems.

A few approaches have studied FL training performance under different hyper-parameters~\cite{flField}. However, they do not consider CompT, TransT, CompL, and TransL together, which is essential from a system's perspective. 
Besides, they almost all use the number of training rounds for comparison, i.e., round-to-accuracy~\cite{fedprox, fedNova20neurips} and convergence rate~\cite{FedAvg17aistats, fedAvgConverge20iclr}. The corresponding observations cannot be directly applied to guide FL system design in our context. This is because clients in FL are heterogeneous in terms of their different amounts of local training data (i.e., \textit{unbalanced} data), computation speeds, and transmission rates, and thus each training round has a different time length, computation cost, and transmission cost. 
For example, selecting more clients in each training round results in a better round-to-accuracy performance~\cite{FedAvg17aistats, fedAvgConverge20iclr}. But it does not necessarily mean a better time-efficiency as the time length of each training round increases with the number of selected clients, nor a better transmission-efficiency as more clients need to transmit model parameters in each training round. 
In addition, it is challenging to tune multiple hyper-parameters in order to achieve diverse training preferences, especially when we need to optimize multiple system aspects. For example, it is unclear how to select hyper-parameters to build an FL training solution that is both CompT and TransL-efficient.



This paper targets a new research problem of optimizing the hyper-parameters for FL from a system perspective. To do so, we formulate the system overheads of FL training and conduct extensive measurements to understand FL training performance.
Our measurement results illustrate how to tune hyper-parameters for simple application scenarios and are the basis of our tuning algorithm. 
To avoid manual hyper-parameter selection, we propose \shortname, an algorithm that automatically tunes FL hyper-parameters during model training, respecting application training preferences. In summary, we make the following contributions:
\begin{enumerate}
    \item To the best of our knowledge, \shortname is the first FL hyper-parameter tuning algorithm that jointly considers applications' training preferences for CompT, TransT, CompL, and CompL. We conduct measurements to understand system training overhead and design an automatic tuning algorithm based on our measurement results. 
    
    \item \shortname is a general framework that applies to various FL hyper-parameters. In this paper, we take the number of participants, the number of training passes, and the model complexity as examples to illustrate FL hyper-parameter tuning. 
    
    \item We extensively evaluate \shortname by various applications/datasets (e.g., command detection, handwriting recognition) and aggregation methods (e.g., FedNova and FedAdagrad). \shortname reduces an average of 8.48\%-26.75\% system overhead compared to using fixed FL hyper-parameters for different applications and aggregation methods. In addition, \shortname reduces some application scenarios by up to 70.51\%, showing that \shortname greatly benefits application scenarios that common hyper-parameters perform poorly. 
    
    \item 
    We release our code at \url{https://github.com/DataSysTech/FedTune} to facilitate the research of FL hyper-parameter tuning.  
\end{enumerate}

This paper is organized as follows. First, we introduce related work in Section \ref{sec:related_work}. Then, we clarify the problem and provide intuitive explanations of our preliminary measurement results in Section \ref{sec:problem}. The details of \shortname algorithm is given in Section \ref{sec:fedtune}. Last, we discuss the limitations/opportunities in Section \ref{sec:discuss} and conclude this paper in Section \ref{sec:conclusion}.




\section{Related Work}
\label{sec:related_work}

\begin{table}[!t]
    \centering
    \centerline{
    \scalebox{0.92}{
    \begin{tabular}{c c | c c}
        Work & Description & Single trial  & System \\
        \toprule
        FTS~\cite{fts20neurips} & optimize client models & \xmark & \xmark \\
        DP-FTS-DE~\cite{dp-fts-de21neurips} & trade-off privacy and utility & \xmark & \xmark \\
        Auto-FedRL~\cite{auto-fedrl22arxiv} & improve model accuracy &  \cmark & \xmark \\
        \cite{rl19arxiv} & improve training robustness & \cmark & \xmark \\
        FedEx~\cite{fedex21neurips} & NAS based framework & \xmark & \xmark \\
        FLoRA~\cite{flora} & NAS based framework & \cmark & \xmark \\
        \shortname (Ours) & a lightweight framework & \cmark & \cmark \\
        \bottomrule
    \end{tabular}
    }
    }
    \vspace{0.05in}
    \caption{Related work on FL hyper-parameter optimization. We tag if (1) the work can run in an online and single trail manner and (2) the work targets system overheads of FL training. }
    \vspace{-0.2in}
    \label{tab:related_work}
\end{table}

\begin{figure*}
\centering 
\centerline{
\subfigure[User distribution]{
    \includegraphics[width=1.7in]{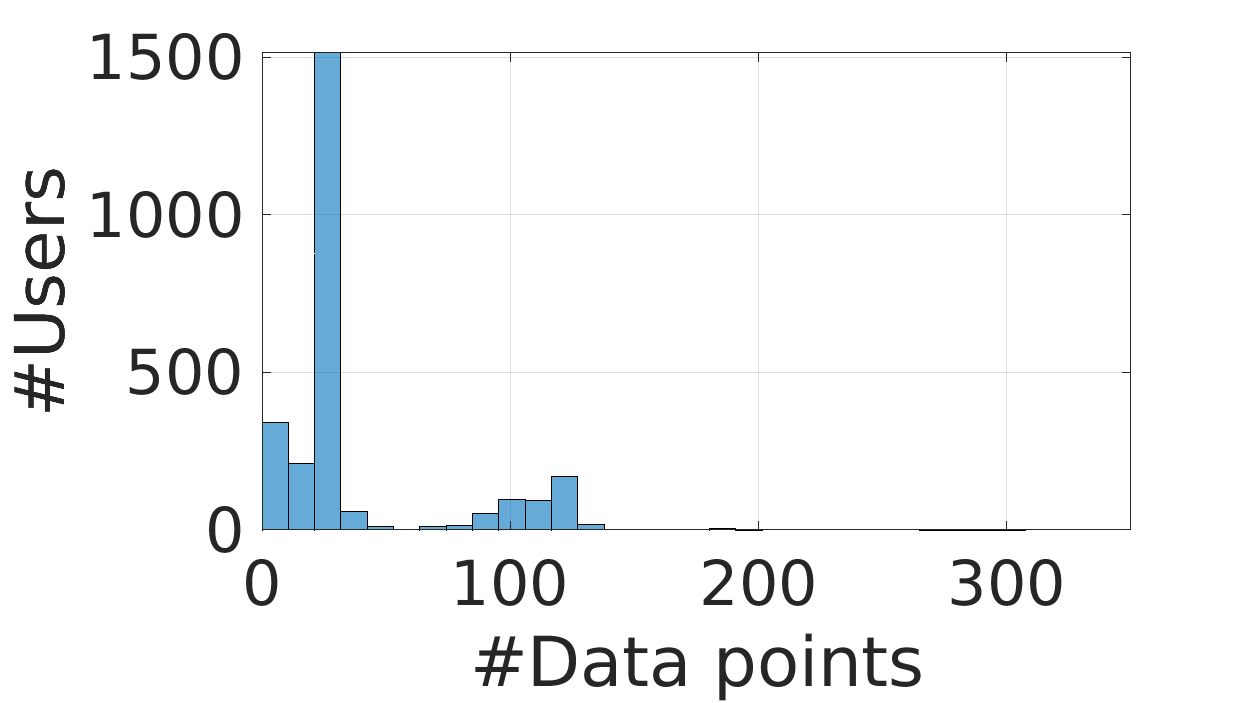}
    \label{fig:dataset_stat}
}
\subfigure[Class distribution]{
    \includegraphics[width=1.7in]{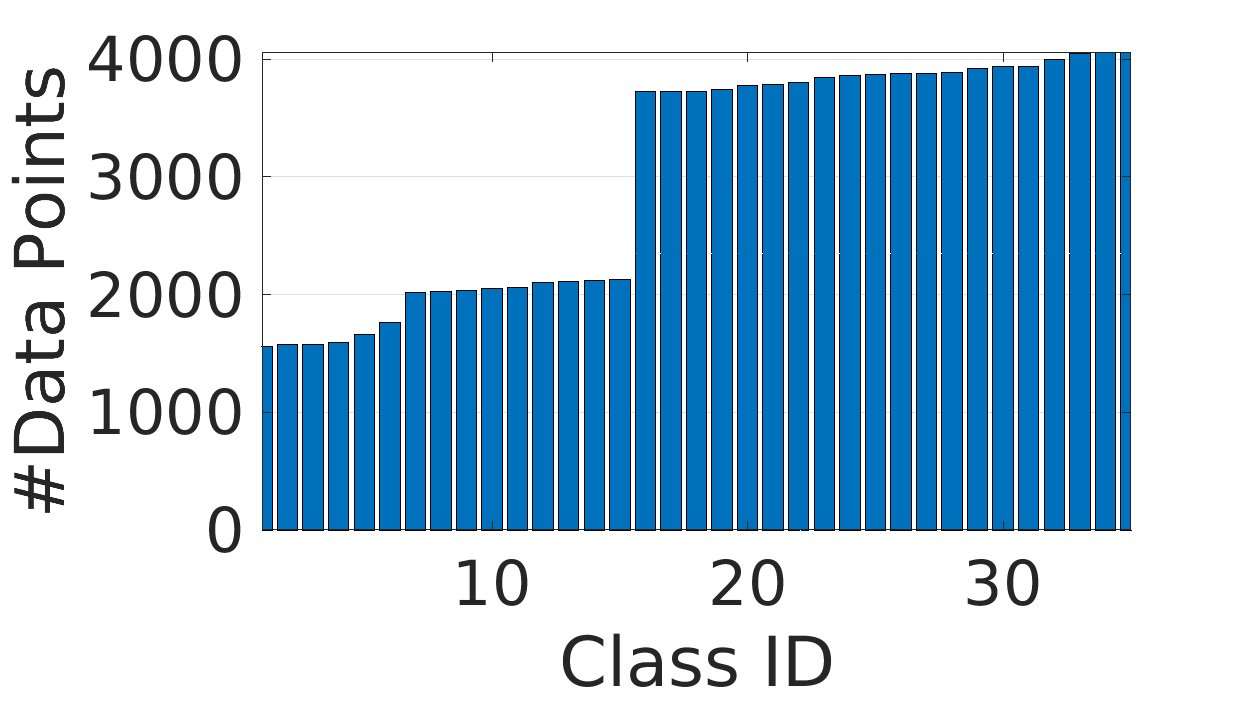}
    \label{fig:dataset_class_distribution}
}
\subfigure[Example spectrograms]{
\raisebox{0.1in}[0pt][0pt]{
    \includegraphics[width=1.8in]{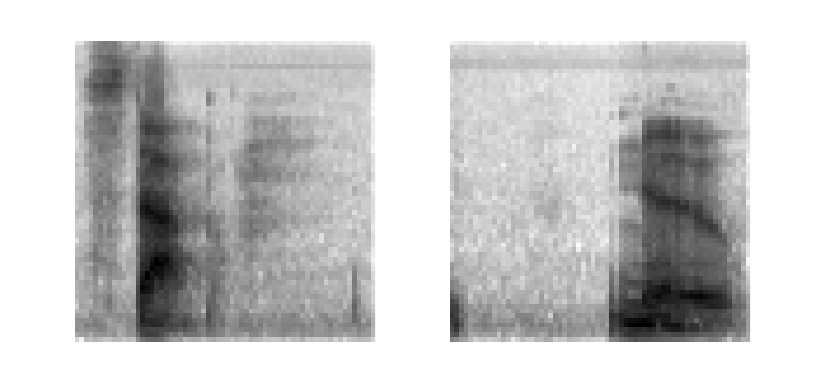}
}
    \label{fig:dataset_spectrogram}
}
}
    \caption{Google speech-to-command dataset used in measurements. }
    \label{fig:dataset}
\end{figure*}

Hyper-Parameter Optimization (HPO) is a field that has been extensively studied~\cite{hpo_survey}. Many classical HPO algorithms, e.g., Bayesian Optimization (BO)~\cite{bayesianOptimization}, successive halving~\cite{halfSuccessing}, and hyperband~\cite{hyperband}, are designed to optimize hyper-parameters of machine learning models. However, they cannot be directly applied to FL because FL has unique hyper-parameters such as the client-side and server-side aggregation methods, and FL is a different training paradigm from centralized machine learning and conventional distributed machine learning.

Designing HPO methods for FL is a new research area. Only a few approaches have touched FL HPO problems. Table~\ref{tab:related_work} lists several representative approaches, where we highlight whether the work can execute in an online and single trial manner and whether the work targets system overhead in FL training.
For example, BO has been integrated with FL to improve different client models~\cite{fts20neurips} and strength client privacy~\cite{dp-fts-de21neurips}. Several approaches apply reinforcement learning to adjust FL hyper-parameters~\cite{auto-fedrl22arxiv, rl19arxiv}, which introduces extra complexity and loss of generality.  FedEx is a general framework to optimize the round-to-accuracy of FL by exploiting the Neural Architecture Search (NAS) techniques of weight-sharing, which improves the baseline by several percentage points~\cite{fedex21neurips}; FLoRA determines the global hyper-parameters by selecting the hyper-parameters that have good performances in local clients~\cite{flora}. Recently, a benchmark suite for federated
hyper-parameter optimization~\cite{fedhpo-b22arxiv} is designed, whose effectiveness remains to be investigated.

For two reasons, existing approaches cannot be directly applied to our scenario of optimizing FL hyper-parameter for different FL training preferences. First, CompT (in seconds), TransL (in seconds), CompL (in FLOPs), and TransL (in bytes) are not comparable with each other. Incorporating training preferences in HPO is not trivial. Second, hyper-parameter tuning needs to be done during the FL training. No ``comeback'' is allowed as the FL model keeps training until its final model accuracy. Otherwise, it will cause significantly more system overhead.


\section{Understanding the Problem}
\label{sec:problem}

\begin{figure*}[h]
    \centering
\subfigure[Accuracy vs training rounds]{
    \includegraphics[width=0.28\textwidth]{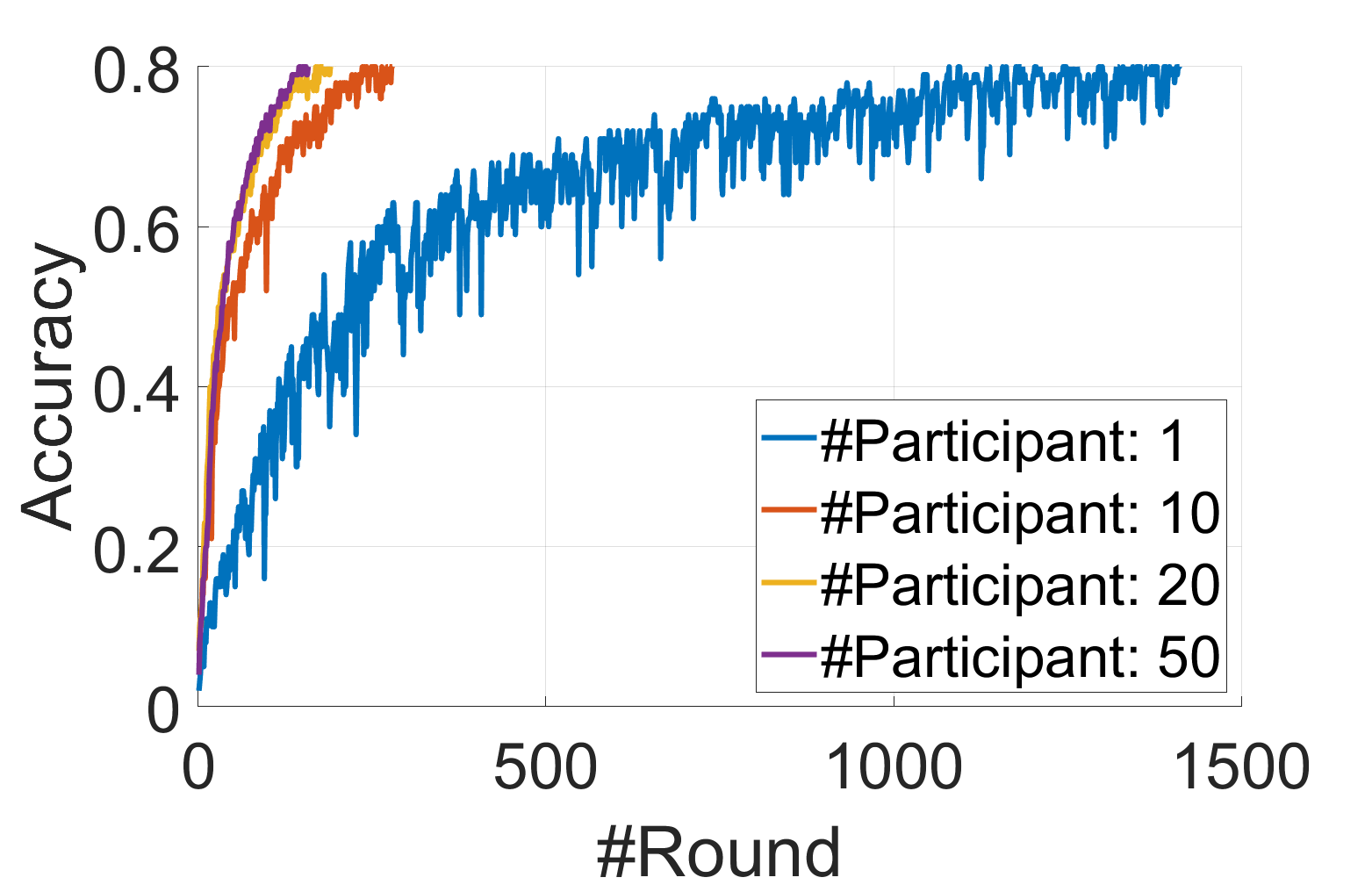}
    \label{fig:accuracy_vs_round}
}
\subfigure[Accuracy vs computation time]{
    \includegraphics[width=0.28\textwidth]{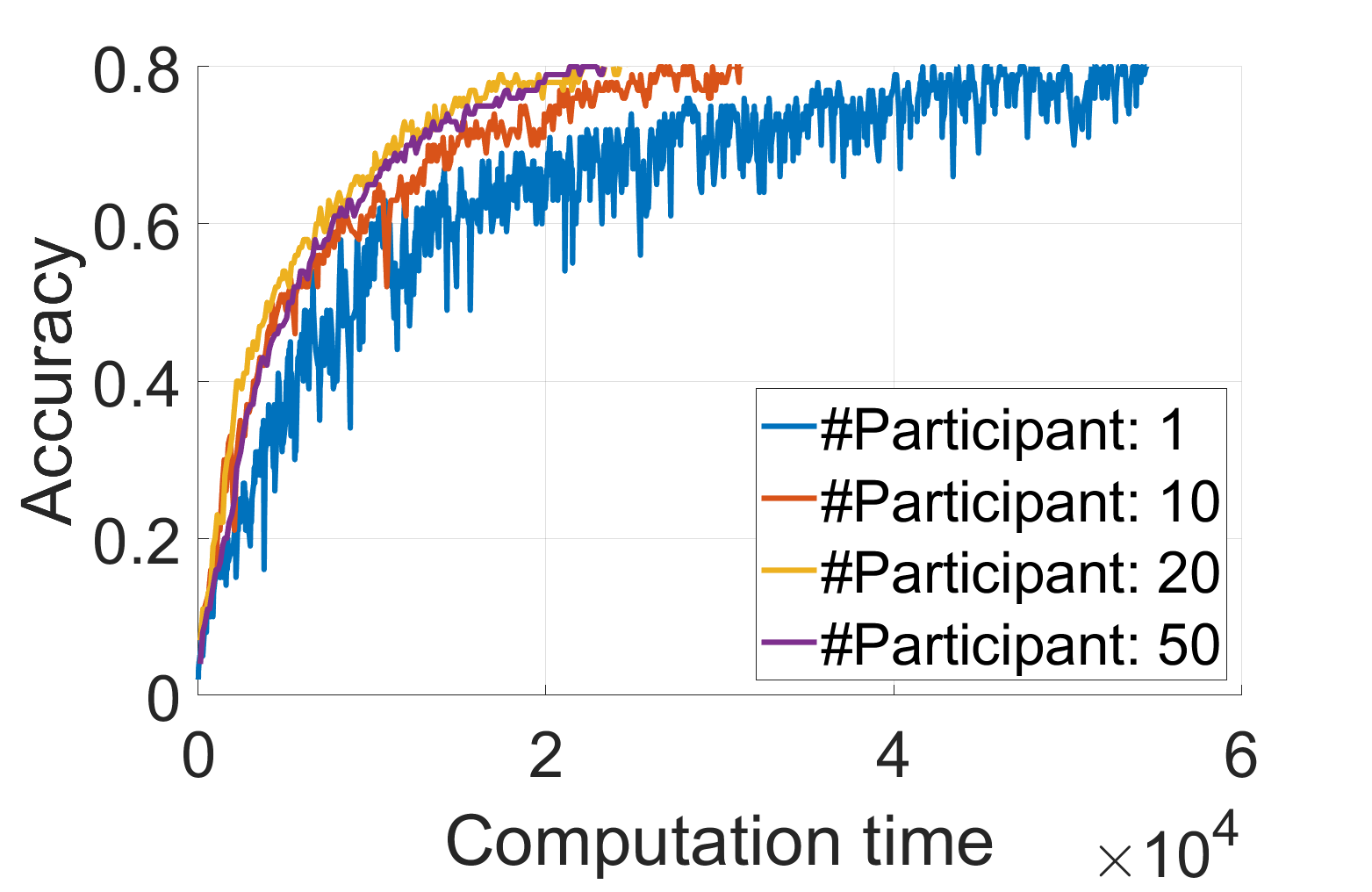}
    \label{fig:accuracy_vs_compT}
}
\subfigure[Time length of round]{
    \includegraphics[width=0.28\textwidth]{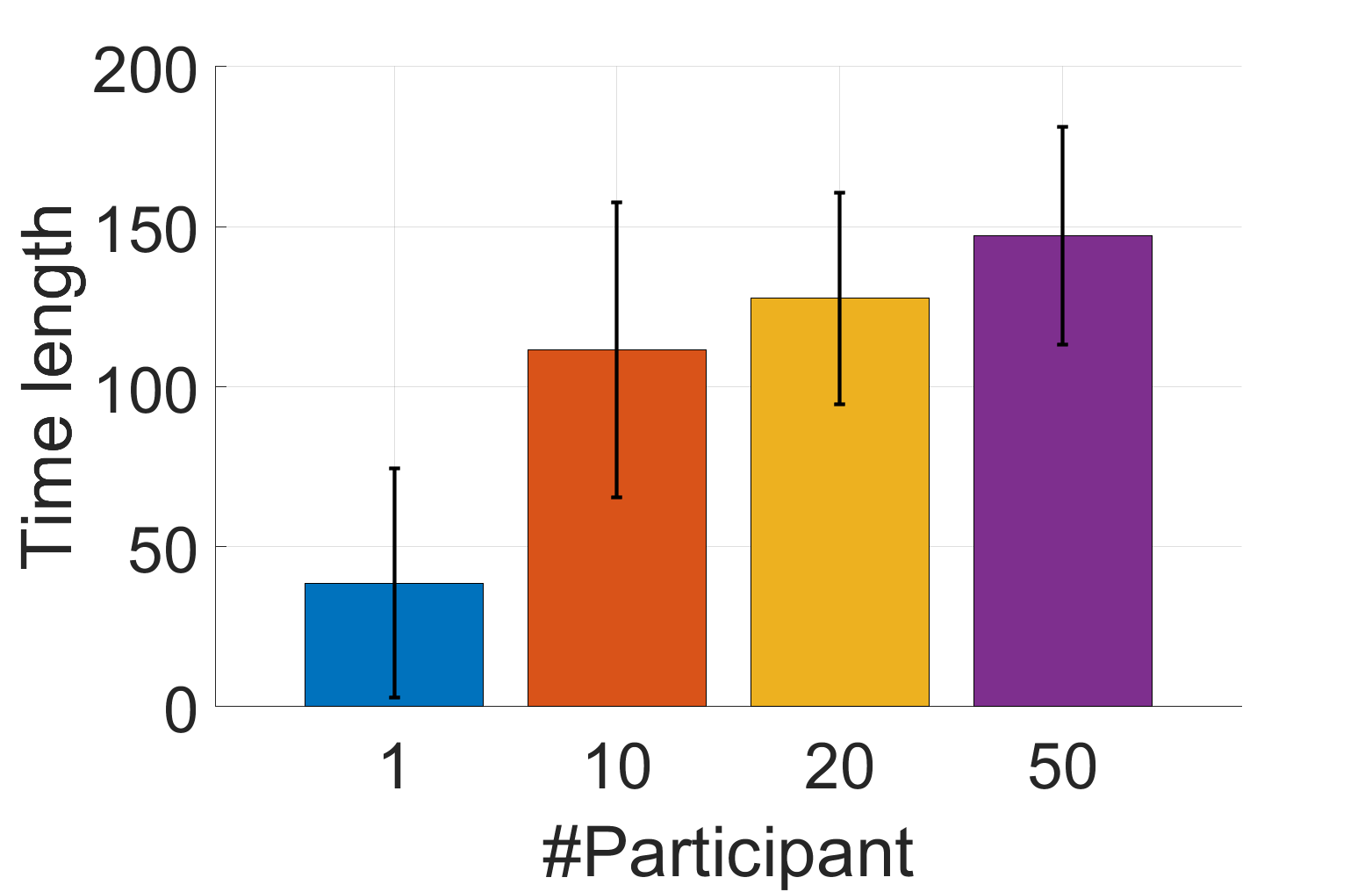}
    \label{fig:round_time_length}
}
\subfigure[Accuracy vs computation load]{
    \includegraphics[width=0.28\textwidth]{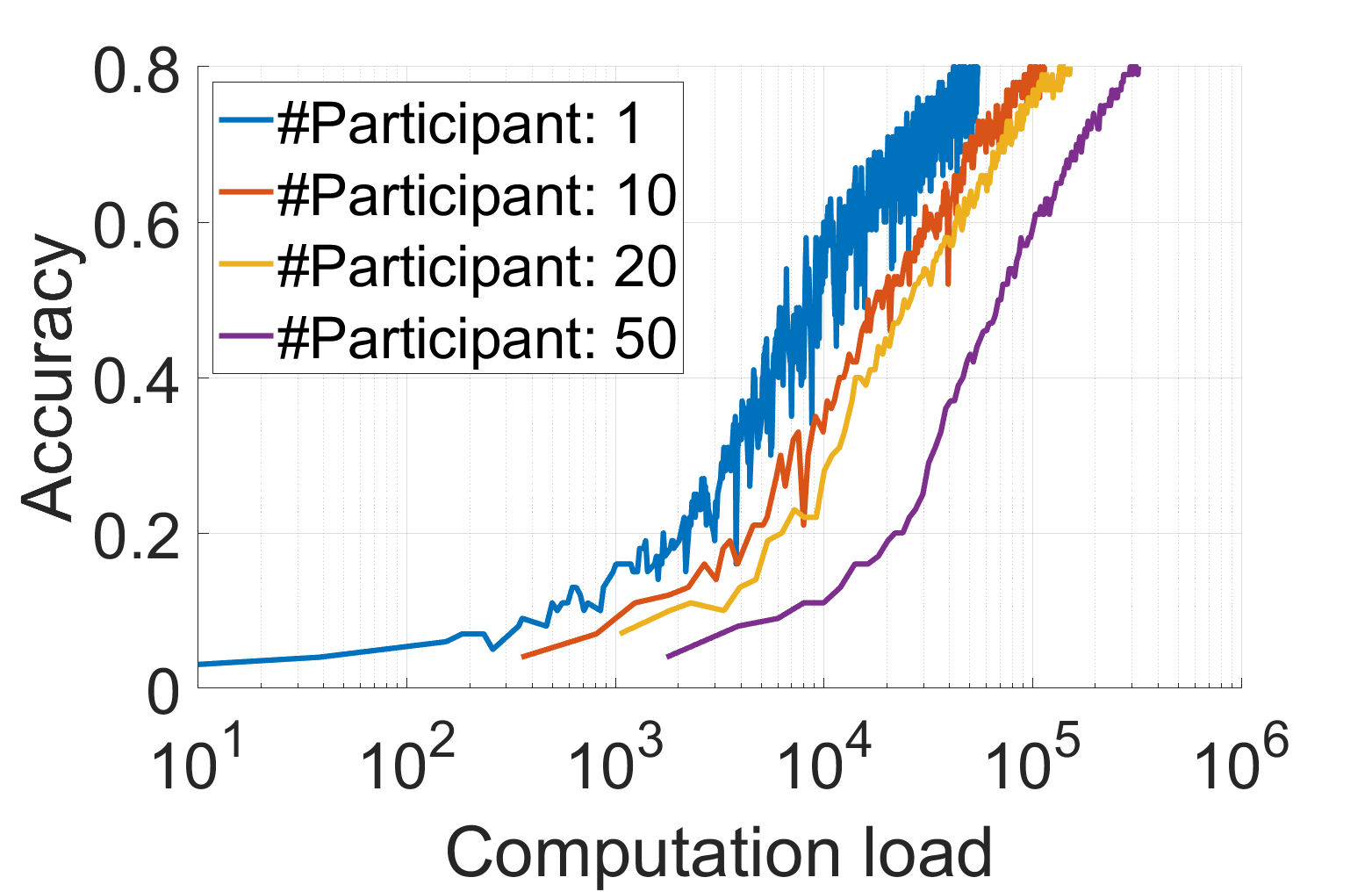}
    \label{fig:accuracy_vs_compL}
}
\subfigure[Accuracy vs transmission time]{
    \includegraphics[width=0.28\textwidth]{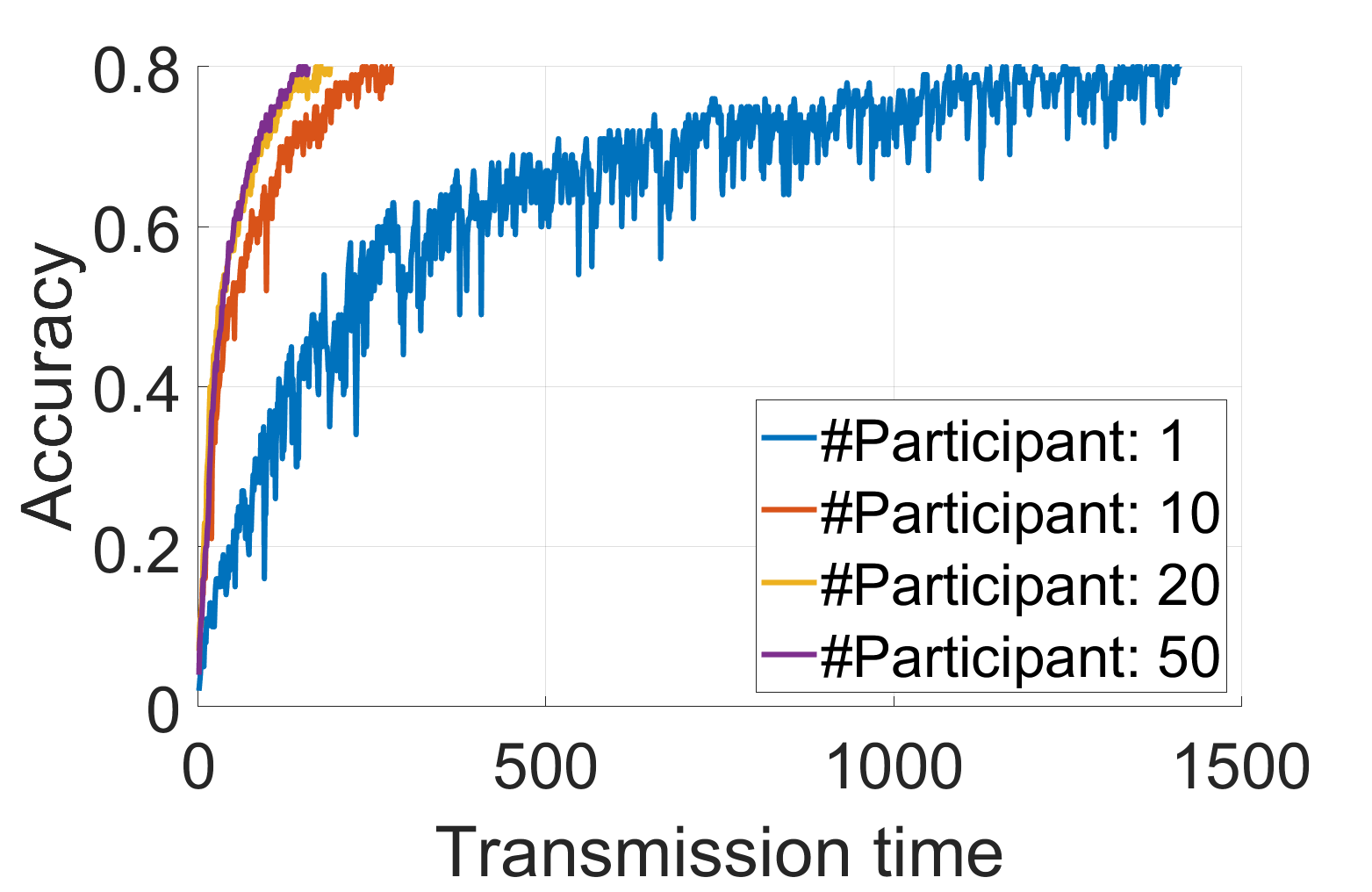}
    \label{fig:accuracy_vs_transT}
}
\subfigure[Accuracy vs transmission load]{
    \includegraphics[width=0.28\textwidth]{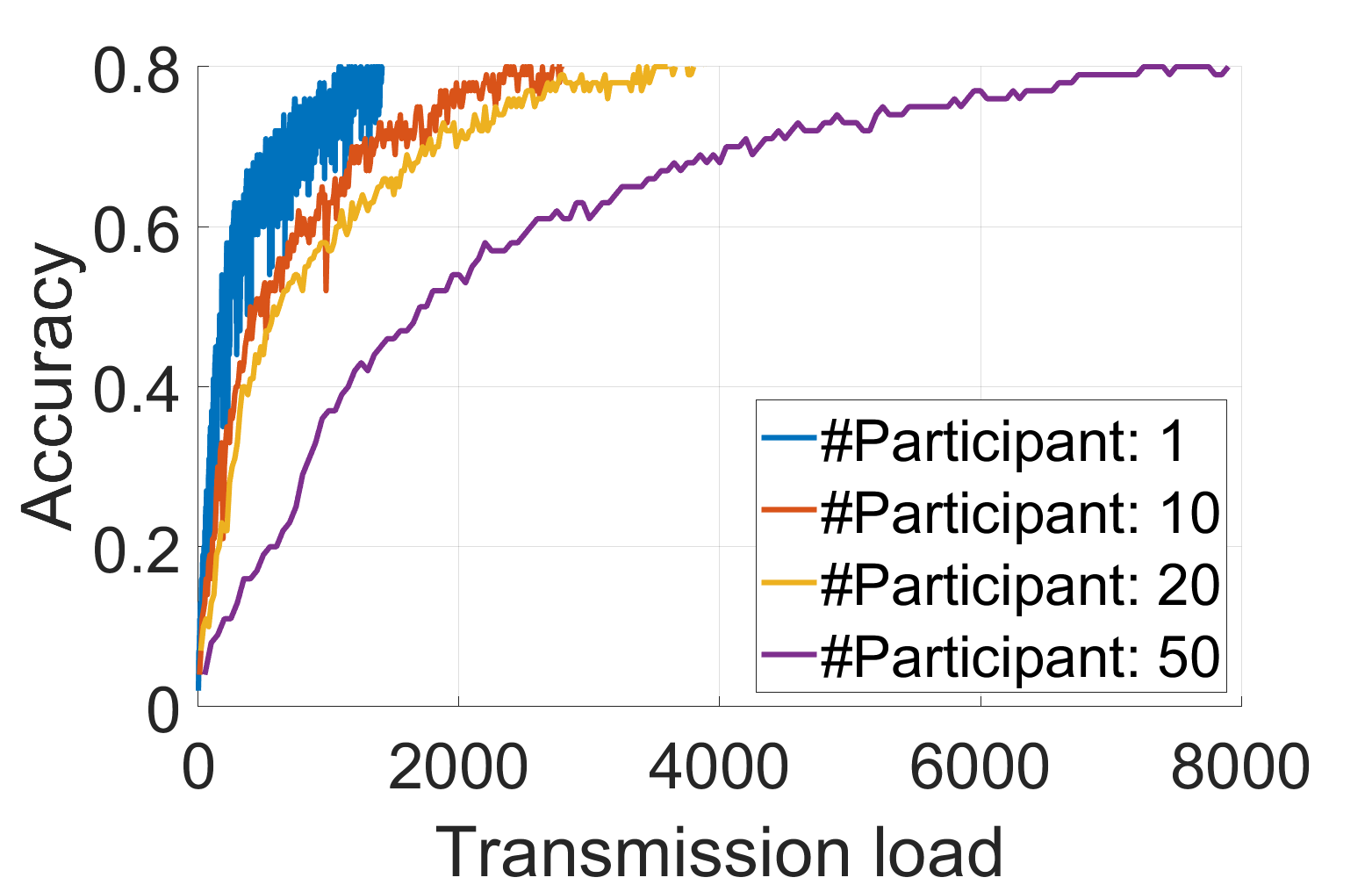}
    \label{fig:accuracy_vs_transL}
}
    \caption{Illustration of FL training when a different number of participants are used in the measurement. }
    \label{fig:appendix_trace_illustration}
\end{figure*}

We first quantify the system overhead of FedAvg to illustrate the problem.  
FedAvg minimizes the following objective 

\begin{equation}
    f(w) = \sum_{k=1}^{K} \frac{n_k}{n} F_k(w) \quad \text{where} \quad F_k(w) = \frac{1}{n_k} \sum_{i \in \mathcal{P}_k} f_i(w)
\end{equation}
where $f_i(w)$ is the loss of the model on data point $(x_i, y_i)$, that is,  $f_i (w) =\ell (x_i, y_i; w)$,  $K$ is the total number of clients, $\mathcal{P}_k$ is the set of indexes of data points on client $k$, with $n_k = |\mathcal{P}_k|$, and $n$ is the total number of data points from all clients, i.e., $n = \sum_{k=1}^K n_k$. Due to the large number of clients in a typical FL application (e.g., millions of clients in the Google Gboard project~\cite{gboard}), a common practice is to randomly select a small fraction of clients in each training round. In the rest of this paper, we refer to the selected clients as participants and denote $M$ as the number of participants in each training round. 
Each participant makes $E$ training passes over its local data in each round before uploading its model parameters to the server for aggregation. Afterward, participants wait to receive an updated global model from the server, and a new training round starts.

\subsection{System Model}

\begin{table}[!t]
    \centering
\scalebox{0.95}{
    \begin{tabular}{c c c c c}
        Model & ResNet-10 & ResNet-18 & ResNet-26 & ResNet-34 \\
        \toprule
        \#BasicBlock & [1, 1, 1, 1] & [2, 2, 2, 2] & [3, 3, 3, 3] & [3, 4, 6, 3] \\
        \#FLOP ($\times 10^6$) & 12.5 & 26.8 & 41.1 & 60.1 \\
        \#Params ($\times 10^3$) & 79.7 & 177.2 & 274.6 & 515.6 \\
        Accuracy & 0.88 & 0.90 & 0.90 & 0.92 \\
        \bottomrule
    \end{tabular}
    }
    \vspace{0.05in}
    \caption{Different models used for the measurement study.}
    \vspace{-0.2in}
    \label{tab:model}
\end{table}


Assume that clients are homogeneous regarding hardware~(e.g., CPU/GPU) and network~(e.g., transmission speeds). Let $b_{k, r}$ indicate whether client $k$ participates at the training round $r$. Then, we have $\sum_{k=1}^{K} b_{k, r} = M$, i.e., each round selects $M$ participants. The number of training rounds to reach the final model accuracy is denoted by $R$, which is unknown \textit{a priori} and varies when different sets of FL hyper-parameters are used in FL training. 
CompT, TransT, CompL, and TransL can be formulated as follows. 

\begin{itemize}
    \item 
    \textbf{Computation Time (CompT)}. If client $k$ is selected in a training round, its training time can be represented by $C_1 \cdot E \cdot n_k$, where $C_1$ is a constant. It is proportional to its number of data points~(i.e., $n_k$) because $n_k$ decides the number of local updates (number of mini-batches) for one epoch, and each local update includes one forward-pass and one backward-pass. 
The computation time of the training round $r$ is determined by the slowest participant and thus is represented by $ C_1 \cdot E \cdot \max_{k=1}^K b_{k, r} \cdot n_k$. In total, the computation time of an FL model training can be formulated as 
\begin{equation}
    CompT = C_1 \cdot E \cdot \sum_{r=1}^{R}\max_{k=1}^K b_{k, r} \cdot n_k  
    \label{equ:compT}
\end{equation}

\item 
\textbf{Transmission Time (TransT)}. Each participant in a training round needs one download and one upload of model parameters from and to the server~\cite{flField}. Thus, the transmission time is the same for all participants in any training round, i.e., a constant $C_2$. The total transmission time is represented by 
\begin{equation}
    TransT = C_2 \cdot R  
    \label{equ:transT}
\end{equation}

\item 
\textbf{Computation Load (CompL)}.
Client $k$ has $C_3 \cdot E \cdot n_k$ computation load if it is selected in a training round, where $C_3$ is a constant. The computation load of the training round $r$ is the summation of each participant's computation load and thus is $C_3 \cdot E \cdot \sum_{k=1}^K b_{k, r} \cdot  n_k$. We can formulate the overall computation load as
\begin{equation}
    CompL = C_3 \cdot E \cdot \sum_{r=1}^R \sum_{k=1}^K b_{k, r} \cdot  n_k
    \label{equ:compL}
\end{equation}

\item 
\textbf{Transmission Load (TransL)}. Since each training round selects $M$ participants, the transmission load for a training round is $C_4 \cdot M$ where $C_4$ is a constant. The total number of training rounds is $R$, and thus, the total transmission load of an FL training is represented by 
\begin{equation}
    TransL = C_4 \cdot R \cdot M
    \label{equ:transL}
\end{equation}

\end{itemize}

As we assume that clients have homogeneous hardware and network, clients have the same $C_1$, $C_2$, $C_3$, and $C_4$. Thus, these constants do not affect the comparison of training overhead under different hyper-parameters when the same model is used. 
In the experiments, we assign the model's number of FLOPs for one input to $C_1$ and $C_3$, and the model's number of parameters to $C_2$ and $C_4$.

\subsection{Measurement Setup}
\label{sect:measurement}

We conduct measurements to study the system overheads when different FL hyper-parameters are used for training.
The measurements aim to help us better understand FL training and are the basis of our automatic tuning algorithm.
We use the Google speech-to-command dataset~\cite{speechToCommandDataset}, which classifies one second’s audio clip into 35 commands such
as yes, no, right, and up. The dataset includes audio clips that are crowd-sourced from 2618 clients.
As officially suggested in~\cite{speechToCommandDataset}, we use 2112 clients’ data for training and the remaining
506 clients’ data for testing. 
Fig.~\ref{fig:dataset_stat} shows the distribution of the number of clients versus the number of data points. It is clear that clients' data are heterogeneous: many clients have only one data point, while others can have up to 316 data points. Fig.~\ref{fig:dataset_class_distribution} plots the histogram of each class's number of data points, which shows that the overall data distribution is unbalanced. The speech-to-command dataset demonstrates the three data properties of FL: massively distributed, unbalanced, and non-IID.
In the measurement study, we investigate the FL training overhead in terms of the following three hyper-parameters.

\begin{itemize}
    \item The number of participants (i.e., $M$). It is well-known that more participants in each training round have a better round-to-accuracy performance~\cite{FedAvg17aistats}. 
    However, the time efficiency is not necessarily better because the time length of each training round also increases with more participants. 
    Besides, it is unclear how computation and communication overheads behaves for different number of participants.
    In the measurement study, we set $M$ to 1, 10, 20, and 50. 
    
    \item The number of training passes (i.e., $E$). Increasing the number of training passes as a method to improve communication efficiency has been adopted in several works, such as FedAvg~\cite{FedAvg17aistats} and FedNova~\cite{fedAvgConverge20iclr}. 
    However, how does the number of training passes affect the time overhead and computation overhead is unclear. 
    In the measurement study, we set $E$ to 0.5, 1, 2, 4, 8, where 0.5 means that only half of each client's local data are used for local training in each round. 
    
    \item Model complexity. We also investigate how the model complexity influences the training overheads if a target accuracy is met. Although it is common knowledge that smaller models have better time and computation performance in other paradigms of model training, we are the first to report the four system overhead versus model complexity in the FL setting. 
    We use ResNet~\cite{resnet} to build different models, as listed in Table \ref{tab:model}.
\end{itemize}

\subsection{Illustration of FL Training}
\label{sec:fl_illustration}

To gain an intuitive understanding of FL training, we plot FL training profiles in Fig.~\ref{fig:appendix_trace_illustration} where a different number of participants $M$ is used on each round. We use ResNet-18 and set the target model accuracy to 0.8. In this measurement, $E$, $C_1$, $C_2$, $C_3$ and $C_4$ are all set to 1, for illustration purpose. Results are normalized to the largest overhead.

As expected, more participants lead to a better accuracy-to-round performance (Fig.~\ref{fig:accuracy_vs_round}). However, if the FL training is stopped too early, it is misleading to conclude that more participants result in higher final model accuracy. Instead, they all reach the same model accuracy as FL hyper-parameters do not invalidate the convergence. 
Correspondingly, more participants have a better accuracy-to-CompT performance, as shown in Fig.~\ref{fig:accuracy_vs_compT}. However, the performance gap between a different number of participants for the accuracy-to-CompT is less significant than the accuracy-to-round performance. This is because the time duration of each training round increases when more participants are selected on each training round (Fig.~\ref{fig:round_time_length}). In this measurement, there are no obvious accuracy-to-CompT difference when 20 and 50 participants are selected. On the other hand, fewer participants lead to better accuracy-to-CompL performance, as shown in Fig.~\ref{fig:accuracy_vs_compL}, which means the effectiveness of each unit computation operation is higher when fewer participants are involved in the FL training. The accuracy-to-TransT (Fig.~\ref{fig:accuracy_vs_transT}) is the same as the accuracy-to-round (Fig.~\ref{fig:accuracy_vs_round}) since we adopt $C_3=1$ for plotting. That is, more participants have a better accuracy-to-TransT performance. Fig.~\ref{fig:accuracy_vs_transL} shows that the accuracy-to-TransL performance has the opposite trend, i.e., more participants result in a worse performance. 

\subsection{Measurement Results}

\begin{figure}[!t]
\centering
\centerline{
\subfigure[Computation Time]{
    \includegraphics[width=1.7in]{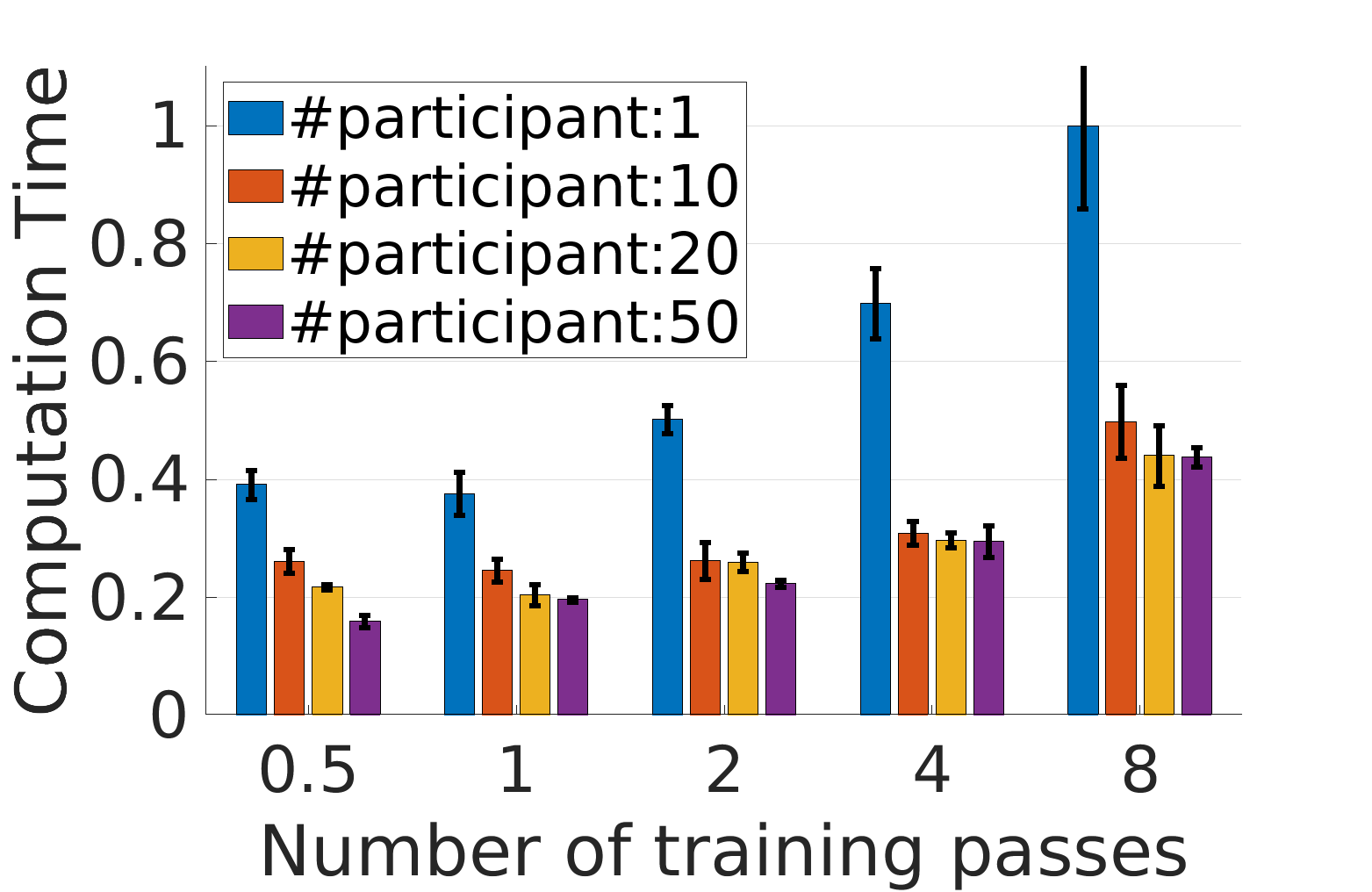}
    \label{fig:measure_M_E_compT}
}
\subfigure[Transmission Time]{
    \includegraphics[width=1.7in]{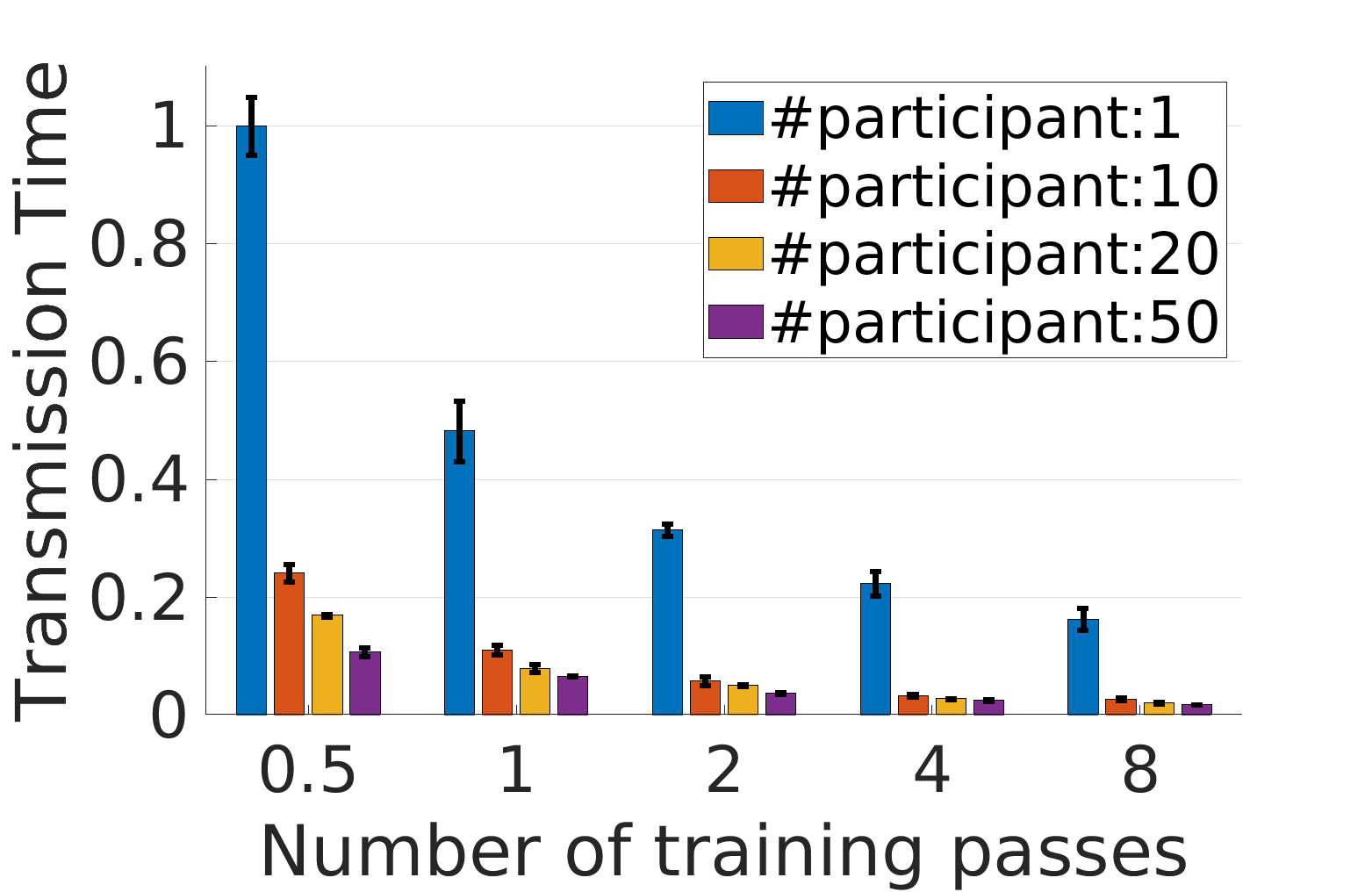}
    \label{fig:measure_M_E_transT}
}
}
\centerline{
\subfigure[Computation Load]{
    \includegraphics[width=1.7in]{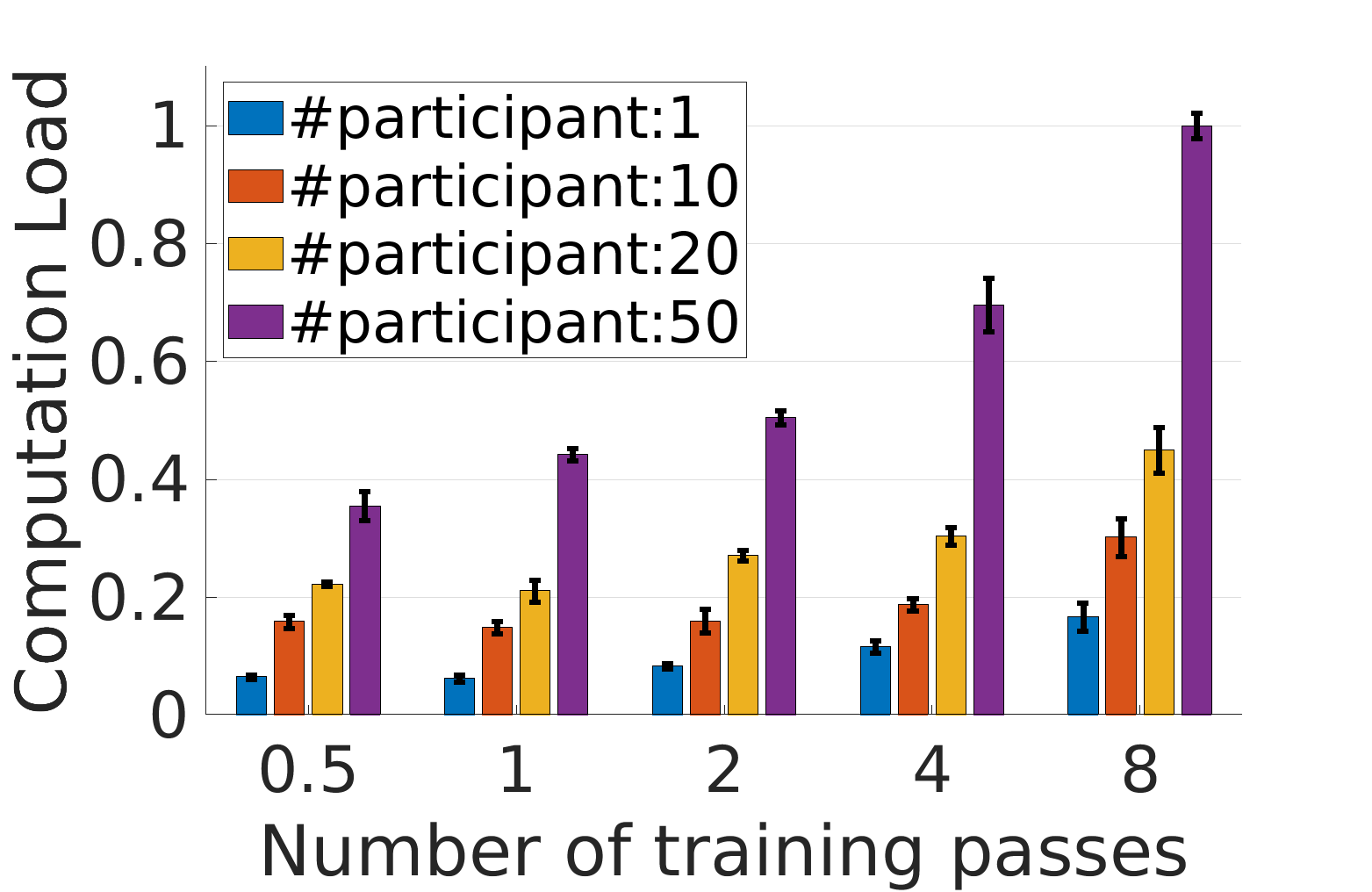}
    \label{fig:measure_M_E_compL}
}
\subfigure[Transmission Load]{
    \includegraphics[width=1.7in]{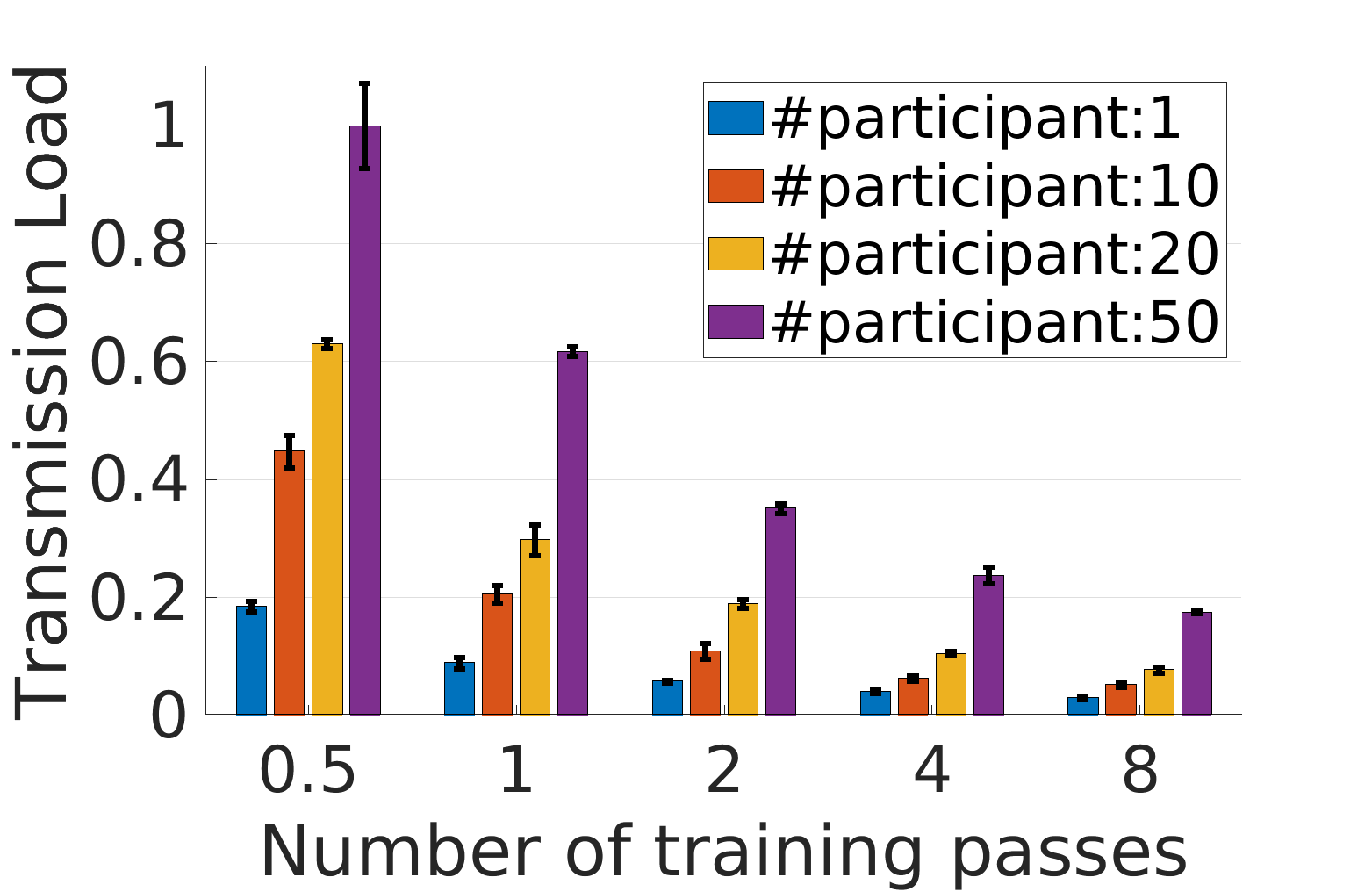}
    \label{fig:measure_M_E_transL}
}
}
    \caption{CompT, TransT, CompL, and TransL when a different number of participants and a different number of training passes are used. The lower the better.}
    \label{fig:measure_M_E}
\end{figure}

\begin{figure}[!t]
    \centering
\centerline{
    \subfigure[Computation time and load]{
    \includegraphics[width=1.7in]{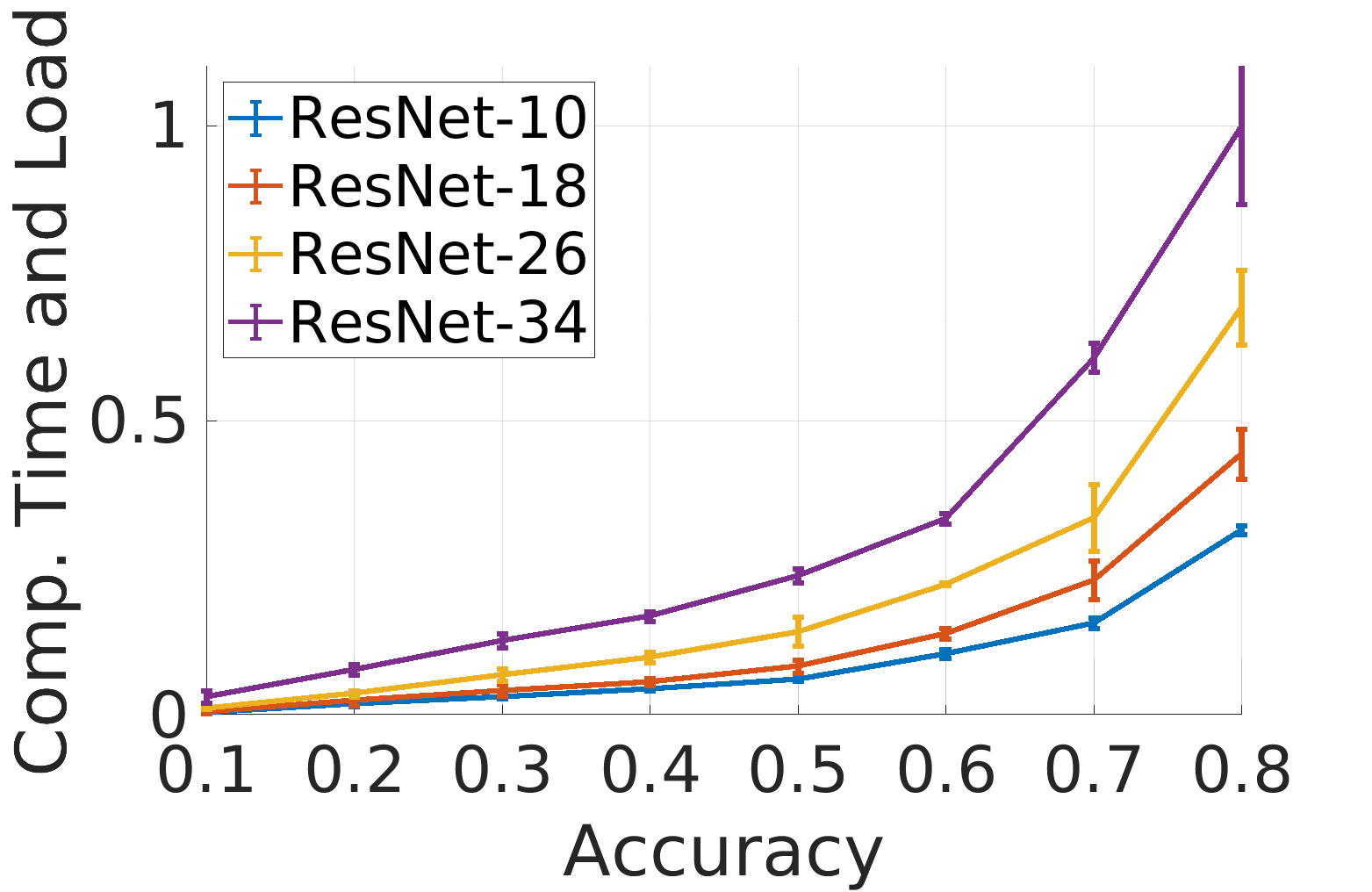}
    \label{fig:model_time_comp}
    }
    \subfigure[Transmission time and load]{
    \includegraphics[width=1.7in]{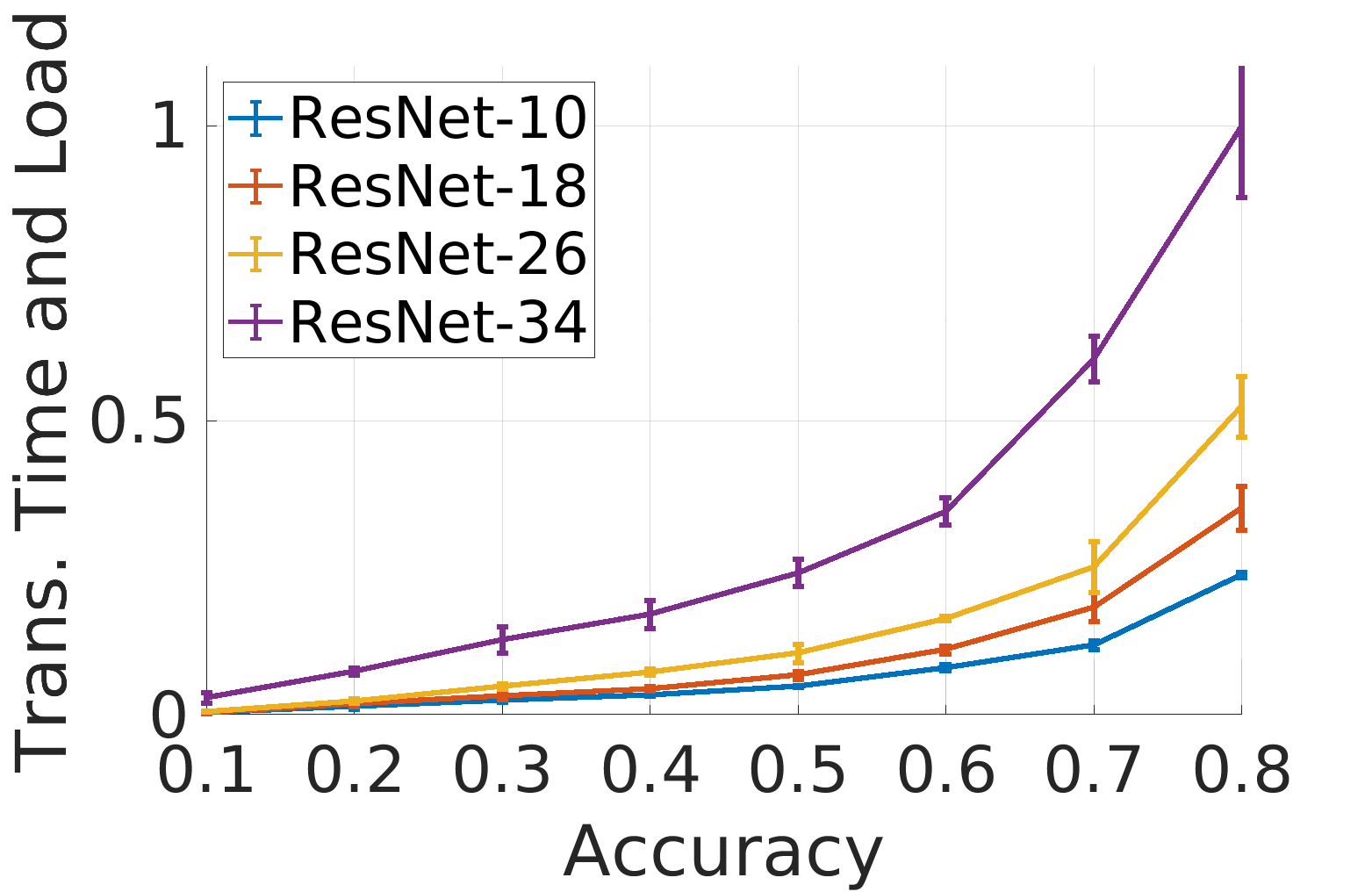}
    \label{fig:model_comm}
    }
}
    \caption{CompT, TransT, CompL, and TransL versus model complexity. The lower the better.}
    \label{fig:model}
\end{figure}

\begin{table}[!t]
\centering
\centerline{
\scalebox{0.95}{
    \begin{tabular}{c | p{0.5in} p{0.5in} c}
      Training aspect & \centering $M$ & \centering $E$ & Model complexity \\\toprule
      CompT & \centering $>$ & \centering $<$ & $<$ \\ 
        CompL & \centering $<$ & \centering $<$ & $<$ \\
      TransT & \centering $>$ & \centering $>$ & $<$ \\ 
      TransL & \centering $<$ & \centering $>$ & $<$ \\
      Model Accuracy & \centering $=$ & \centering $=$ & $>$ \\\bottomrule
    \end{tabular}
}
}
\vspace{0.05in}
    \caption{System overheads versus the number of participants $M$, the number of training passes $E$, and model complexity. `$<$', `$=$', and `$>$' means the smaller the better, does not matter, and the larger the better, respectively.}
    \label{tab:summary}
\end{table}

\begin{figure*}[!t]
\vspace{0.8in}
    \centering
\centerline{
\subfigure[Number of participants $M$]{
\raisebox{0.1in}[0pt][0pt]{
    \includegraphics[width=0.34\textwidth]{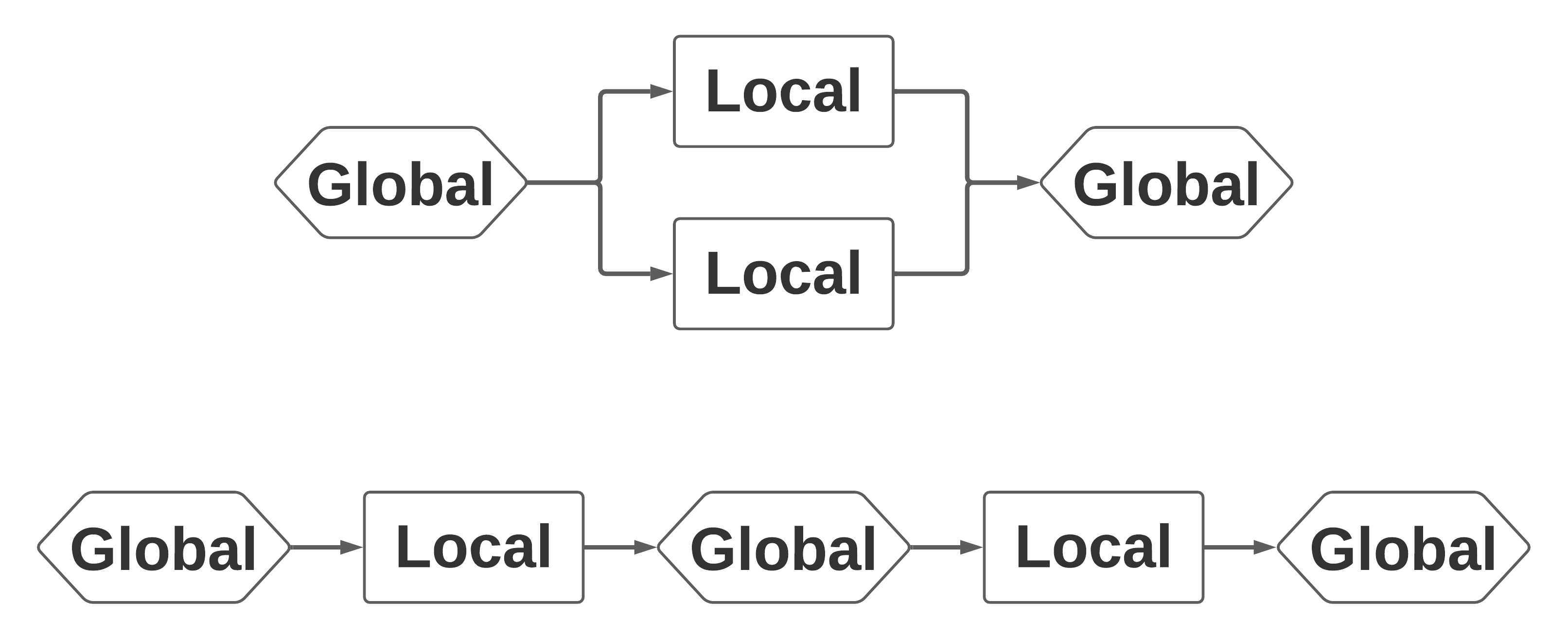}
    \label{fig:heuristic_M}
}
}
\subfigure[Number of training passes $E$]{
\raisebox{0.1in}[0pt][0pt]{
    \includegraphics[width=0.36\textwidth]{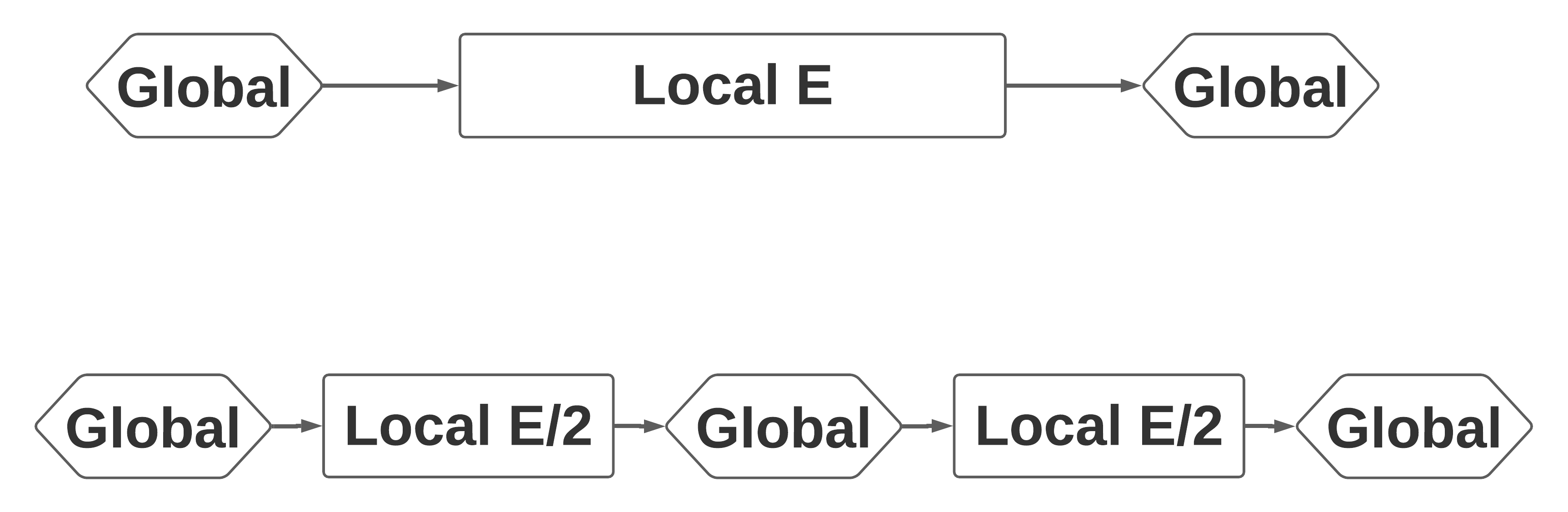}
    \label{fig:heuristic_E}
}
}
\subfigure[Model complexity]{
\raisebox{0.1in}[0pt][0pt]{
    \includegraphics[width=0.22\textwidth]{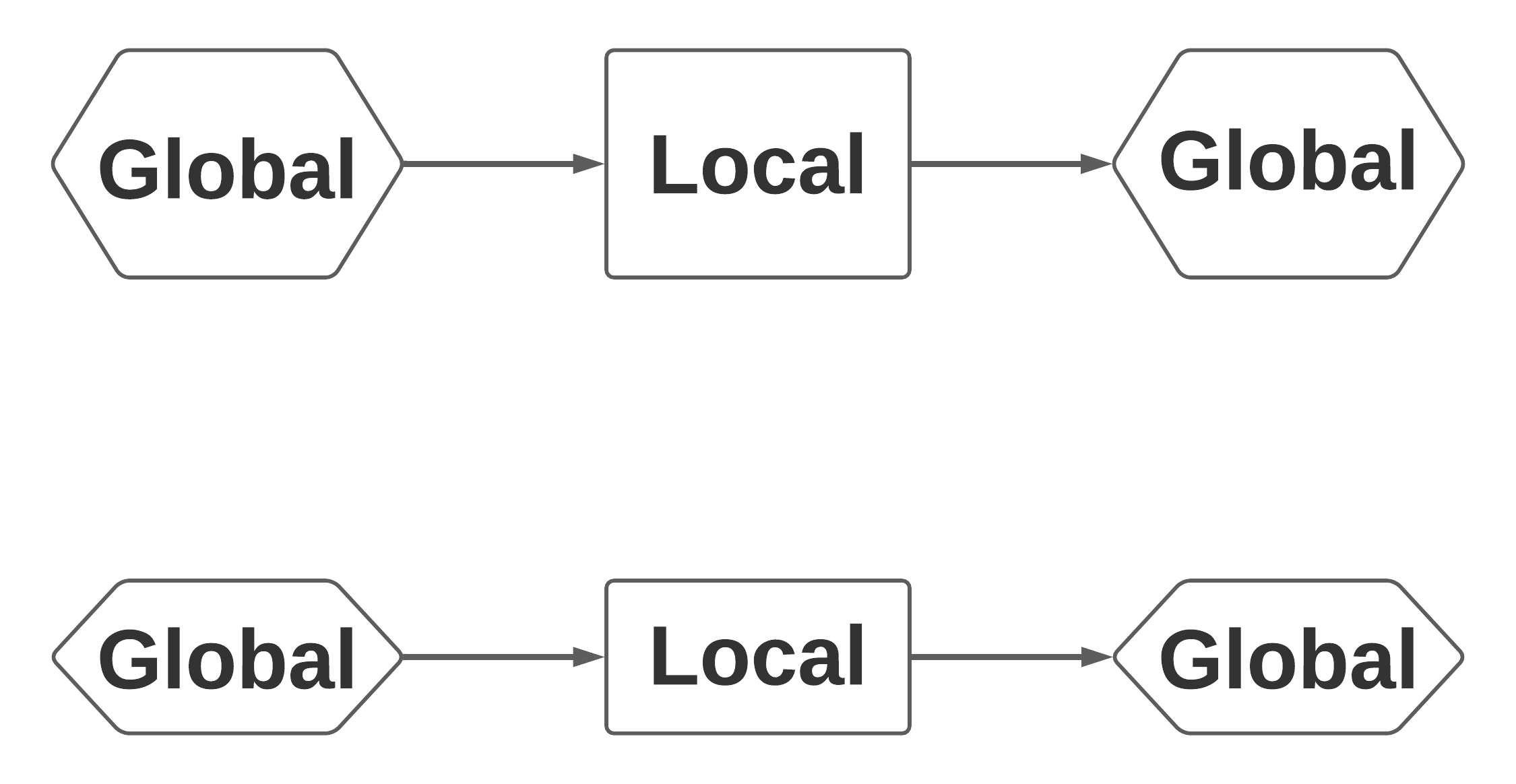}
    \label{fig:heuristic_model}
}
}
}
    \caption{Intuitive explanation of our measurement results. (a) The number of participants. The bottom scheme is better regarding CompL and TransL. (b) The number of training passes. The bottom scheme is better regarding CompT and CompL. (c) Model complexity. The bottom scheme is better for CompT, CompL, TransT, and TransL.}
    \label{fig:heuristic}
\end{figure*}

In Section \ref{sec:fl_illustration}, we illustrate CompT, CompL, TransT, and TransL performance when a different number of participants are adopted in FL training. This section presents results from more extensive measurements for the number of participants $M$, the number of training passes $E$, and the model complexity. The results are averaged over three runs. 

\begin{itemize}
    \item 
    \textbf{Computation Time (CompT)}.
Fig.~\ref{fig:measure_M_E_compT} compares CompT for a different number of participants $M$ and a different number of training passes $E$. In the experiments, we use ResNet-18 and normalize their overheads. As we can see, more participants lead to smaller CompT, i.e.; it takes a shorter time to converge. However, the difference is insignificant among 10, 20, and 50 participants, especially when the number of training passes is large. 
In addition, we can see that larger $E$ has worse CompT. 
There is no apparent difference between $E=0.5$ and $E=1$ though. In a nutshell, the common knowledge of more participants is faster for FL model training is valid. However, the gain of more participants is insignificant when the number of participants is moderate. In addition, it is preferred to adopt a small number of training passes to achieve good time efficiency.

\item 
\textbf{Transmission Time (TransT)}. Fig.~\ref{fig:measure_M_E_transT} plots TransT, which clearly shows that TransT favors larger $M$ and $E$. Since TransT is dependent on the number of training rounds $R$ (Eq.~(\ref{equ:transT})), it is equivalent to the metric of round-to-accuracy. Our measurement result is consistent with common knowledge (e.g., \cite{fedNova20neurips}) that more participants and more training passes have a better round-to-accuracy performance. We can also observe that when $M$ is small, e.g., 1, TransL is much worse than the other cases.

\item 
\textbf{Computation Load (CompL)}.
Fig.~\ref{fig:measure_M_E_compL} shows CompL. We make the following observations: (1) More participants result in worse CompL. The results indicate that the gain of faster model convergence from more participants does not compensate for the higher computation costs introduced by more participants. (2) CompL is increased when more training passes are used. This is probably because that larger $E$ diverges the model training~\cite{fedprox}, and thus, the data utility per unit of computation cost is reduced. 

\item 
\textbf{Transmission Load (TransL)}.
Fig.~\ref{fig:measure_M_E_transL} illustrates TransL. As shown, more participants greatly increase TransL. This is because more participants can only weakly reduce the number of training rounds $R$~\cite{fedAvgConverge20iclr}, however, in each round, the number of transmissions increases linearly with the number of participants. Regarding the number of training passes,  larger $E$ reduces the total number of training rounds $R$ and thus has better TransL. On the other hand, the gain of larger $E$ diminishes. The results are consistent with the analysis of \cite{fedAvgConverge20iclr} that $R$ is hyperbolic with $E$ (the turning point happens around 100-1000 in their experiments).

\item 
\textbf{Model Complexity.}
Table \ref{tab:model} tabulates the models for comparing training overheads versus model complexity. In this experiment, we select one participant ($M=1$) to train one pass ($E=1$) on each training round. 
Fig.~\ref{fig:model} shows the normalized CompT, TransT, CompL, and TransL for different models. The x-axis is the target model accuracy, and the y-axis is the corresponding overhead to reach that model accuracy. Since only one client and one training pass are used on each round, CompT and CompL have the same normalized comparison, and so are TransT and TransL. The results show that smaller models are better in terms of all training aspects. 
In addition, it is interesting to note that heavier models have higher increase rates of overheads versus model accuracy. This means that model selection is especially essential for high model accuracy applications.

\end{itemize}

\subsection{Summary of System Overhead}

Based on our measurement study, we summarize systems overheads versus FL hyper-parameters in Table \ref{tab:summary}.  As we can see, CompT, TransT, CompL, and TransL conflict with each other in selecting the optimal $M$ and $E$.
Regarding model complexity, smaller models have better system overheads if the model accuracy is satisfied. Please note that Table \ref{tab:summary} is consistent with existing work (e.g., \cite{fedNova20neurips}), but is more comprehensive. 
Thus, Table \ref{tab:summary} is also valid for other datasets and ML models. 
\subsection{Intuitive Explanation of Measurement Results}

We present intuitive explanation of the measurement results to help FL practitioners better understand FL hyper-parameters.  
Fig.~\ref{fig:heuristic} visualizes the intuition.

\begin{itemize}
    \item \emph{The number of participants $M$}. In  Fig.~\ref{fig:heuristic_M}, the top scheme and the bottom scheme have the same computation cost (two local training) and communication cost (four transmissions). However, the bottom scheme has a better overall CompL and TransL at the expense of degraded CompT and TransT. The bottom scheme is better probably because the clients in the bottom scheme always work on the updated global model, whereas the clients in the top scheme work on the same global model. In other words, narrow-and-deep FL schemes have better computation load and transmission load than wide-and-shallow FL schemes. 
    
    \item \emph{The number of training passes $E$}. In Fig.~\ref{fig:heuristic_E}, both the top scheme and the bottom scheme take $E$ total training passes. However, the bottom scheme has better CompT and CompL at the expense of higher communication costs. The results indicate that the usefulness per local update decreases with the number of local updates. Therefore, for computation-sensitive FL applications, large $E$ should be avoided.

    \item \emph{Model complexity}. Fig.~\ref{fig:heuristic_model} shows that a smaller model has better CompT, TransT, CompL, and TransL than a heavier model, as long as the accuracy requirement is met.  In addition, our results in Fig.~\ref{fig:model} indicate that if the target model accuracy is low, then using a heavy model does not introduce significantly more overhead than a lightweight model. However, if the goal is to achieve a high-accuracy model, carefully selecting a model complexity is essential, as over-large models cause significantly more system overhead.

\end{itemize}

\section{\shortname: Automatic Tuning of Federated Learning
Hyper-Parameters from A System Perspective}
\label{sec:fedtune}

Due to the conflicting FL hyper-parameters for CompT, TransT, CompL, and TransL, practitioners bear the burden of selecting a set of FL hyper-parameters, which if not chosen well may lead to poor system performance. For example, it is unclear how to set FL hyper-parameters that are both CompT and TransT friendly, since these two system aspects prefer different number of training passes. 
We propose \shortname to adjust FL hyper-parameters during the FL training automatically.
\shortname considers the application's preference for
CompT, TransT, CompL, and TransL, denoted by $\alpha$, $\beta$, $\gamma$, and $\delta$, respectively. We have $\alpha + \beta + \gamma + \delta  = 1$. For example, $\alpha = 0.6$, $\beta = 0.2$, $\gamma = 0.1$, and $\delta = 0.1$ represent that the application cares most about CompT, somewhat about TransT, and the least on CompL and TransL.

\subsection{Problem Formulation}

\begin{algorithm}[!t]
\small
\SetAlgoLined
\textbf{Input: } \\
$\alpha$, $\beta$, $\gamma$, $\delta$: training preference for CompT, TransT, CompL, and TransL\\

\textbf{Begin} \\
$\epsilon$: minimum improvement of model accuracy\\
$D$: penalty factor \\
$S_{prv}$, $S_{cur}$: previous and current sets of hyper-parameters \\
$a_{prv}$, $a_{cur}$: previous and current model accuracy \\
$t_{prv}$, $q_{prv}$, $z_{prv}$, and $v_{prv}$: CompT, TransT, CompL, and TransL under $S_{prv}$ \\
$t_{cur}$, $q_{cur}$, $z_{cur}$ and $v_{cur}$: CompT, TransT, CompL, and TransL under under $S_{cur}$ \\
\For{training round $r = 1, 2, ...$ } {
Train FL model for one round as usual \\
Update $a_{cur}$, $t_{cur}$, $q_{cur}$, $z_{cur}$, and $v_{cur}$ \\
\If{$a_{cur} - a_{prv} > \epsilon$} { 
\tcp{Normalize CompT, TransT, CompL, and TransL}
$t_{cur} = t_{cur} / (a_{cur} - a_{prv}), ...$ \\
\tcp{Calculate comparison function in Eq.~(\ref{equ:compare})}
$I(S_{prv}, S_{cur}) = \alpha \times \frac{t_{cur} - t_{prv}}{t_{prv}} + ...$ \\
\eIf{$M_{cur}>M_{prv}$}{
Update $\theta_{t-1}$ and $\theta_{q-1}$ \\
\If{$I(S_{prv}, S_{cur}) > 0$}{
$\theta_{z-1} = D \times \theta_{z-1}$ \\
$\theta_{v-1} = D \times \theta_{v-1}$
}
}{
Make changes to $\theta_{t-1}$, $\theta_{q-1}$, $\theta_{z-1}$, and $\theta_{v-1}$ \\
}
Make changes to $\eta_{t-1}$, $\eta_{q-1}$, $\eta_{z-1}$, and $\eta_{v-1}$ \\
\tcp{Calculate $\Delta M$ in Eq.~(\ref{equ:Delta_M})}
$\Delta M = \frac{(+1) \times \alpha \times \eta_{t-1} \times |t_{cur} - t_{prv}| }{t_{cur}} + ...$ \\
\tcp{calculate $\Delta E$ in Eq.~(\ref{equ:Delta_E})}
$\Delta E = \frac{(-1) \times \alpha \times \zeta_{t-1} \times |t_{cur} - t_{prv}| }{t_{cur}} + ...$ \\

\eIf{$\Delta M > 0$}{
    $M_{nxt} = M_{cur} + 1$
}{
    $M_{nxt} = M_{cur} -1$
}
\eIf{$\Delta E > 0$}{
    $E_{nxt} = E_{cur} + 1$
}{
    $E_{nxt} = E_{cur} -1$
}
$S_{nxt} = \{M_{nxt}, E_{nxt} \}$ \\
$a_{prv} = a_{cur}$, $t_{prv} = t_{cur}$, $q_{prv} = q_{cur}$, $z_{prv} = z_{cur}$, $v_{prv} = v_{cur}$, $S_{prv} = S_{cur}$, $S_{cur} = S_{nxt}$ \\
Change FL hyper-parameters according to $S_{nxt}$ \\
}
}
\caption{\shortname update.}
\label{alg:auto_tune}
\end{algorithm}

For two sets of FL hyper-parameters $S_1$ and $S_2$, \shortname defines the comparison function $I(S_1, S_2)$ as
\begin{equation}
\begin{aligned}
    I(S_1, S_2) = \alpha \times \frac{t_2 - t_1}{t_1} + \beta \times \frac{q_2 - q_1}{q_1} \\ + \gamma \times \frac{z_2 - z_1}{z_1} + \delta \times \frac{v_2 - v_1}{v_1}
    \label{equ:compare}
\end{aligned}
\end{equation}
where 
$t_1$ and $t_2$ are CompT for $S_1$ and $S_2$ when achieving the same model accuracy. Correspondingly, $q_1$ and $q_2$ are TransT, $z_1$ and $z_2$ are CompL, and $v_1$ and $v_2$ are TransL. If $I(S_1, S_2) < 0$, then $S_2$ is better than $S_1$.  A set of hyper-parameters is better than another set if the weighted improvement of some training aspects (e.g., CompT and CompL) is higher than the weighted degradation, if any, of the remaining training aspects (e.g., TransT and TransL). The weights are training preferences on CompT, TransT, CompL, and TransL.

However, the training overhead for different sets of FL hyper-parameters are unknown \textit{a priori}. As a result,  directly identifying the optimal hyper-parameters before FL training is impossible. Instead, we propose an iterative method to optimize the next set of hyper-parameters. Given the current set of hyper-parameters $S_{cur}$, the goal is to find a set of hyper-parameters $S_{nxt}$ that improves the training performance the most, that is, minimizes the following objective function:
\begin{equation}
\begin{aligned}
    G(S_{nxt}) =  \alpha \times \frac{t_{nxt} - t_{cur}}{t_{cur}} + \beta \times \frac{q_{nxt} - q_{cur}}{q_{cur}} \\ + \gamma \times \frac{z_{nxt} - z_{cur}}{z_{cur}}  + \delta \times \frac{v_{nxt} - v_{cur}}{v_{cur}}
\end{aligned}
\end{equation}
where $t_{cur}$, $q_{cur}$, $z_{cur}$, and $v_{cur}$ are CompT, TransT, CompL, and TransL under the current hyper-parameters $S_{cur}$;  $t_{nxt}$, $q_{nxt}$, $z_{nxt}$, and $v_{nxt}$ are CompT, TransT, CompL, and TransL for the next hyper-parameters $S_{nxt}$. We focus on the number of participants $M$ and the number of training passes $E$, since training overhead is monotonous with model complexity. Therefore, we need to optimize $S_{nxt} = \{M_{nxt}, E_{nxt}\}$. 

\subsection{$S_{nxt}$ Optimization}

\begin{table*}[h]
    \centering
    \centerline{
\scalebox{0.98}{
\begin{tabular}{c c c c | c c c c | c c | c}
$\alpha$ & $\beta$ & $\gamma$  & $\delta$ & CompT ($10^{12}$) & TransT ($10^6$) & CompL ($10^{12}$) & TransL ($10^6$) & Final M & Final E & Overall \\\toprule 
- & - & - & - & 0.94 (0.01) & 11.61 (0.10) & 5.97 (0.04) & 232.24 (1.99) & 20 & 20 & - \\
1.0 & 0.0 & 0.0 & 0.0 & 0.42 (0.02) & 50.19 (2.57) & 4.57 (0.22) & 2418.71 (240.91) & 57.33 (4.50) & 1.00 (0.00) & +55.23\% (2.22\%) \\
0.0 & 1.0 & 0.0 & 0.0 & 1.34 (0.22) & 7.68 (1.12) & 14.99 (2.73) & 289.82 (46.98) & 48.00 (2.16) & 48.00 (2.16) & +33.87\% (9.67\%) \\
0.0 & 0.0 & 1.0 & 0.0 & 1.02 (0.10) & 615.98 (97.52) & 1.76 (0.16) & 672.21 (91.62) & 1.00 (0.00) & 1.00 (0.00) & +70.51\% (2.75\%) \\
0.0 & 0.0 & 0.0 & 1.0 & 2.18 (0.47) & 35.47 (7.51) & 3.30 (0.22) & 76.47 (1.68) & 1.00 (0.00) & 46.67 (3.30) & +67.07\% (0.72\%) \\
0.5 & 0.5 & 0.0 & 0.0 & 0.82 (0.13) & 9.17 (1.26) & 9.13 (1.66) & 347.11 (54.31) & 47.33 (2.05) & 21.33 (4.78) & +16.97\% (9.68\%) \\
0.5 & 0.0 & 0.5 & 0.0 & 0.48 (0.04) & 81.42 (9.83) & 3.23 (0.14) & 1875.99 (155.21) & 25.00 (1.63) & 1.00 (0.00) & +47.57\% (3.43\%) \\
0.5 & 0.0 & 0.0 & 0.5 & 0.79 (0.10) & 11.59 (0.55) & 5.04 (0.89) & 241.86 (68.65) & 22.33 (5.79) & 15.67 (4.50) & +5.82\% (11.28\%) \\
0.0 & 0.5 & 0.5 & 0.0 & 0.83 (0.03) & 10.66 (0.15) & 5.16 (0.31) & 207.79 (6.08) & 21.00 (1.41) & 21.00 (1.41) & +10.87\% (2.83\%) \\
0.0 & 0.5 & 0.0 & 0.5 & 1.54 (0.16) & 11.48 (3.83) & 9.59 (3.52) & 190.52 (61.53) & 19.67 (14.82) & 49.00 (0.00) & +9.55\% (7.08\%) \\
0.0 & 0.0 & 0.5 & 0.5 & 1.69 (0.26) & 50.14 (8.21) & 2.70 (0.26) & 93.21 (8.48) & 1.00 (0.00) & 23.33 (2.49) & +57.32\% (3.76\%) \\
0.33 & 0.33 & 0.33 & 0.0 & 0.82 (0.07) & 11.59 (1.01) & 5.65 (0.27) & 255.35 (9.65) & 22.33 (2.62) & 15.67 (1.25) & +6.09\% (6.67\%) \\
0.33 & 0.33 & 0.0 & 0.33 & 1.06 (0.08) & 10.07 (0.90) & 8.10 (0.34) & 247.54 (29.18) & 26.33 (2.05) & 27.00 (2.16) & -1.93\% (7.40\%) \\
0.33 & 0.0 & 0.33 & 0.33 & 0.91 (0.19) & 18.23 (5.83) & 4.15 (1.13) & 229.26 (63.40) & 12.00 (1.41) & 14.00 (5.72) & +11.66\% (11.76\%) \\
0.0 & 0.33 & 0.33 & 0.33 & 1.13 (0.13) & 16.16 (3.36) & 4.51 (0.59) & 169.93 (25.84) & 9.00 (5.35) & 23.00 (4.55) & +3.99\% (6.19\%) \\
0.25 & 0.25 & 0.25 & 0.25 & 0.91 (0.10) & 9.73 (1.81) & 6.19 (0.76) & 207.34 (3.34) & 23.33 (5.44) & 22.67 (3.30) & +6.51\% (6.13\%) \\\bottomrule
\end{tabular}
}
    }
    \vspace{0.05in}
    \caption{Performance of \shortname for the speech-to-command dataset when FedAdagrad is used for aggregation. \\`$+$' is improvement and `$-$' is degradation. Standard deviation in parentheses. }
    \label{tab:eva_speech_to_command}
\end{table*}

To find the optimal $S_{nxt}$, we take the derivatives of $G(S_{nxt})$ over $M$ and $E$, obtaining
\begin{equation}
\begin{aligned}
    \Delta M = \frac{\partial G(S_{nxt})}{\partial M} =  \frac{\alpha}{t_{cur}} \times \frac{\partial t_{nxt}}{\partial M} + \frac{\beta}{q_{cur}} \times \frac{\partial q_{nxt}}{\partial M} \\ + \frac{\gamma}{z_{cur}} \times \frac{\partial z_{nxt}}{\partial M} + \frac{\delta}{v_{cur}} \times \frac{\partial v_{nxt}}{\partial M}
\end{aligned}
\end{equation}
\begin{equation}
\begin{aligned}
    \Delta E = \frac{\partial G(S_{nxt})}{\partial E} =  \frac{\alpha}{t_{cur}} \times \frac{\partial t_{nxt}}{\partial E} + \frac{\beta}{q_{cur}} \times \frac{\partial q_{nxt}}{\partial E} \\ + \frac{\gamma}{z_{cur}} \times \frac{\partial z_{nxt}}{\partial E} + \frac{\delta}{v_{cur}} \times \frac{\partial v_{nxt}}{\partial E}
\end{aligned}
\end{equation}

We illustrate how to approximate $\Delta M$. The process for $\Delta E$ is similar.
Considering that each step makes a small adjustment of $M$, $\partial t_{nxt} / \partial M$ can be represented by $(+1) \times |t_{nxt} - t_{cur}|$, where $(+1)$ means CompT prefers larger $M$ according to Table~\ref{tab:summary}. To estimate $| t_{nxt} - t_{cur} |$, we apply a linear function $\eta_{t-1} \times | t_{cur} - t_{prv}|$ where $\eta_{t-1} = \frac{|t_{cur} - t_{prv}|}{|t_{prv} - t_{prvprv}|}$ ($t_{prvprv}$ is the CompT at two steps before). Thus, $\eta_{t-1}$ represents the slope of the linear function. Similarly, we have $\eta_{q-1}$,  $\eta_{z-1}$, $\eta_{v-1}$ for TransT, CompL, and TransL when calculating their derivatives over $M$.  
As a result, $\Delta M$ can be approximated as
\begin{equation}
\begin{aligned}
\Delta M = \frac{(+1) \times \alpha \times \eta_{t-1} \times |t_{cur} - t_{prv}| }{t_{cur}} \\ + \frac{(+1) \times \beta \times \eta_{q-1} \times |q_{cur} - q_{prv}| }{q_{cur}} \\ + \frac{(-1) \times \gamma \times \eta_{z-1} \times |z_{cur} - z_{prv}| }{z_{cur}} \\ + \frac{(-1) \times \delta \times \eta_{v-1} \times |v_{cur} - v_{prv}| }{v_{cur}}
\end{aligned}
\label{equ:Delta_M}
\end{equation}
Similarly, we can calculate $\Delta E$ as 
\begin{equation}
\begin{aligned}
\Delta E = \frac{(-1) \times \alpha \times \zeta_{t-1} \times |t_{cur} - t_{prv}| }{t_{cur}} \\ + \frac{(+1) \times \beta \times \zeta_{q-1} \times |q_{cur} - q_{prv}| }{q_{cur}} \\ + \frac{(-1) \times \gamma \times \zeta_{z-1} \times |z_{cur} - z_{prv}| }{z_{cur}} \\ + \frac{(+1) \times \delta \times \zeta_{v-1} \times |v_{cur} - v_{prv}| }{v_{cur}}
\end{aligned}
\label{equ:Delta_E}
\end{equation}
where $\zeta_{t-1}$, $\zeta_{q-1}$, $\zeta_{z-1}$, and $\zeta_{v-1}$ are the parameters for calculating the derivatives of CompT, TransT, CompL, and TransL over $E$. 


\subsection{Decision Making and Parameter Update}

Algorithm \ref{alg:auto_tune} shows the updates in \shortname. \shortname is activated when the model accuracy is improved by at least $\epsilon$ (line 13). 
Then, it normalizes current overheads (line 14) and calculates the comparison function of the previous hyper-parameters $S_{prv}$ and the current hyper-parameters $S_{cur}$ (line 15). Afterward, \shortname updates the parameters, i.e., four $\eta$ and four $\zeta$ (line 16-25). Next, it computes $\Delta M$ and $\Delta E$, and determines the next $M$ and $E$ based on the signs of $\Delta M$ and $\Delta E$ (line 26-27). Specifically, $M_{nxt} = M_{cur} + 1$ if $\Delta M > 0$, otherwise, $M_{nxt} = M_{cur} - 1$. Likewise, \shortname increases $E_{nxt}$ by one if $\Delta E > 0$; else \shortname decreases $E_{nxt}$ by one.  The FL training continues using the new hyper-parameters.  It is clear that \shortname is lightweight and its computational burden is negligible to the FL training: it only requires dozens of multiplication and addition calculations.

\shortname automatically updates $\eta_{t-1}$, $\eta_{q-1}$,  $\eta_{z-1}$, $\eta_{v-1}$, $\zeta_{t-1}$, $\zeta_{q-1}$,  $\zeta_{z-1}$, and $\zeta_{v-1}$ during FL training. At each step, \shortname updates the parameters that favor the current decision. For example, if $M_{cur}$ is larger than $M_{prv}$, \shortname updates $\eta_{t-1}$ and $\eta_{q-1}$ as CompT and TransT prefer larger $M$ (line 17); otherwise, \shortname updates $\eta_{z-1}$ and $\eta_{v-1}$. 

Furthermore, \shortname incorporates a penalty mechanism to mitigate bad decisions. Given the previous hyper-parameters $S_{prv}$ and the current hyper-parameters $S_{cur}$, \shortname calculates the comparison function $I(S_{prv}, S_{cur})$ (line 15). A bad decision occurs if the sign of $I(S_{prv}, S_{cur})$ is positive (line 18). In this case, \shortname multiplies the parameters that are against the current decision by a constant penalty factor, denoted by $D$ ($D \geq 1$). For example, if $I(S_{prv}, S_{cur}) > 0$ and $M_{cur} > M_{prv}$, \shortname updates $\eta_{t-1}$ and $\eta_{q-1}$ as explained before, but also multiplies $\eta_{z-1}$ and $\eta_{v-1}$ by $D$ (line 19-20).

\section{Evaluation}

\begin{figure*}[!t]
    \centering
\centerline{
\subfigure[(1-0-0-0)]{
    \includegraphics[width=0.18\textwidth]{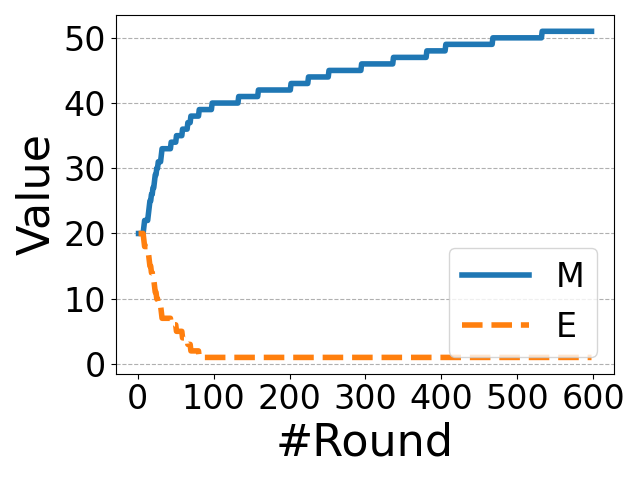}
}
\subfigure[(0-1-0-0)]{
    \includegraphics[width=0.18\textwidth]{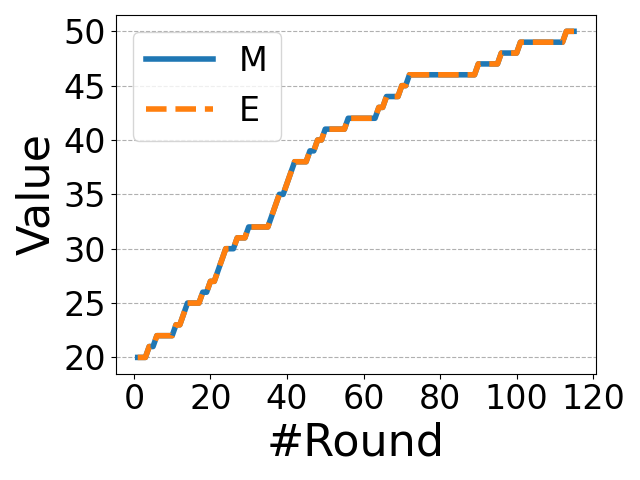}
}
\subfigure[(0-0-1-0)]{
    \includegraphics[width=0.18\textwidth]{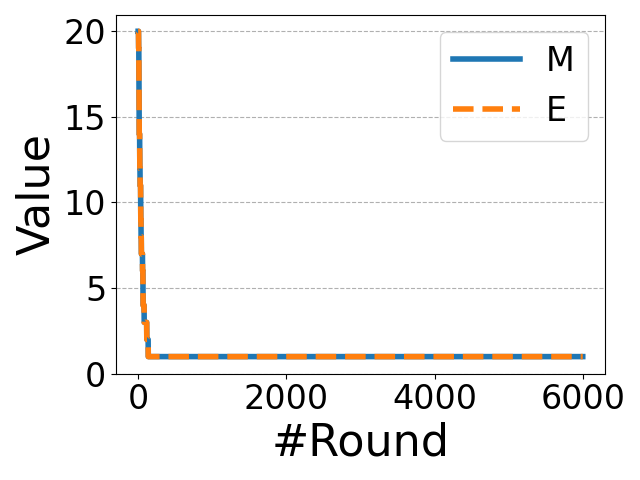}
}
\subfigure[(0-0-0-1)]{
    \includegraphics[width=0.18\textwidth]{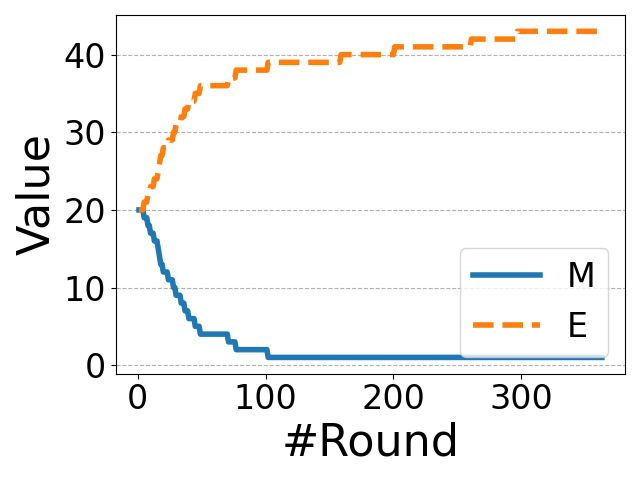}
}
\subfigure[(0.5-0.5-0-0)]{
    \includegraphics[width=0.18\textwidth]{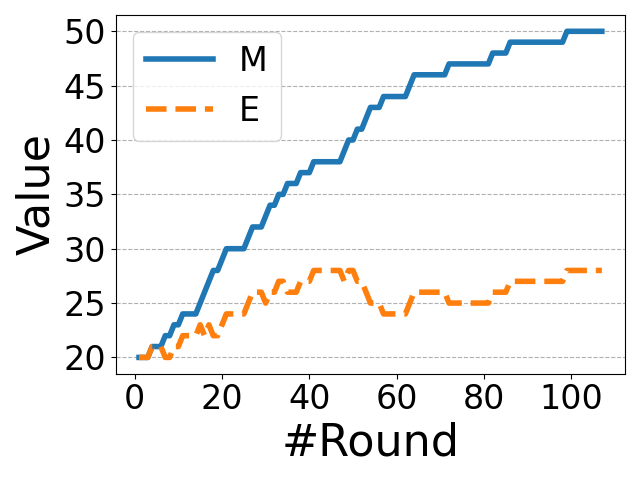}
}
}
\centerline{
\subfigure[(0.5-0-0.5-0)]{
    \includegraphics[width=0.18\textwidth]{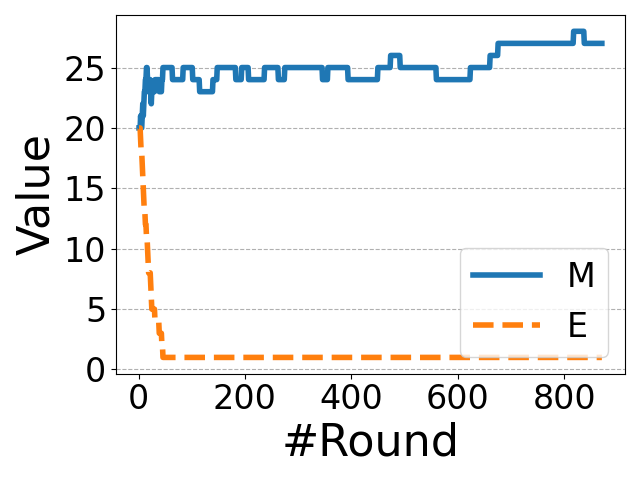}
}
\subfigure[(0.5-0-0-0.5)]{
    \includegraphics[width=0.18\textwidth]{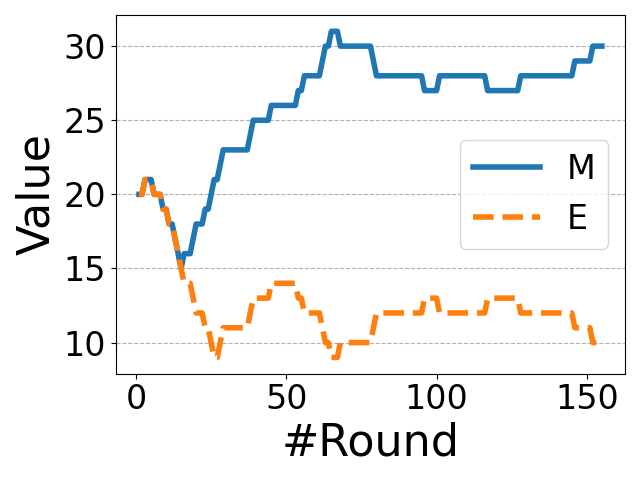}
}
\subfigure[(0-0.5-0.5-0)]{
    \includegraphics[width=0.18\textwidth]{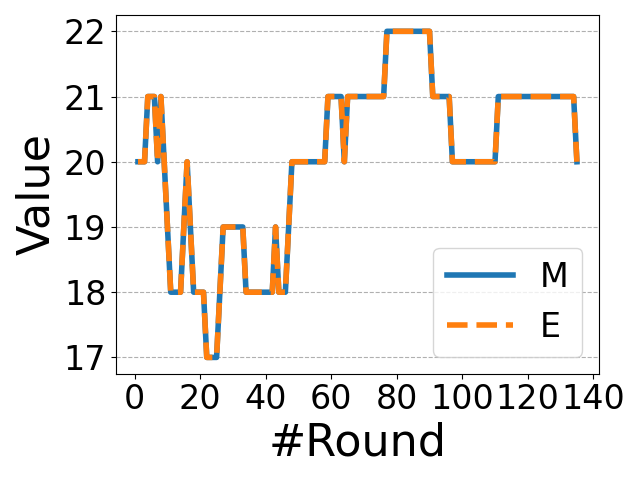}
}
\subfigure[(0-0.5-0-0.5)]{
    \includegraphics[width=0.18\textwidth]{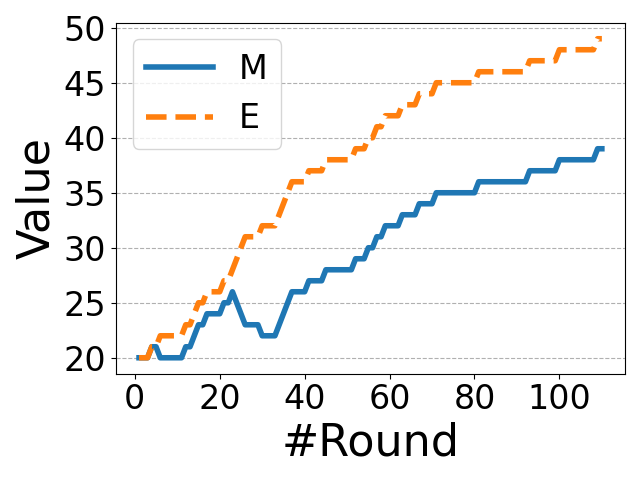}
}
\subfigure[(0-0-0.5-0.5)]{
    \includegraphics[width=0.18\textwidth]{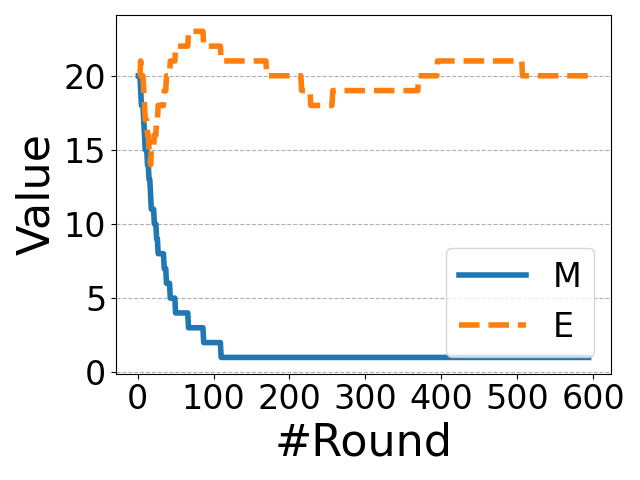}
}
}
\centerline{
\subfigure[(0.33-0.33-0.33-0)]{
    \includegraphics[width=0.18\textwidth]{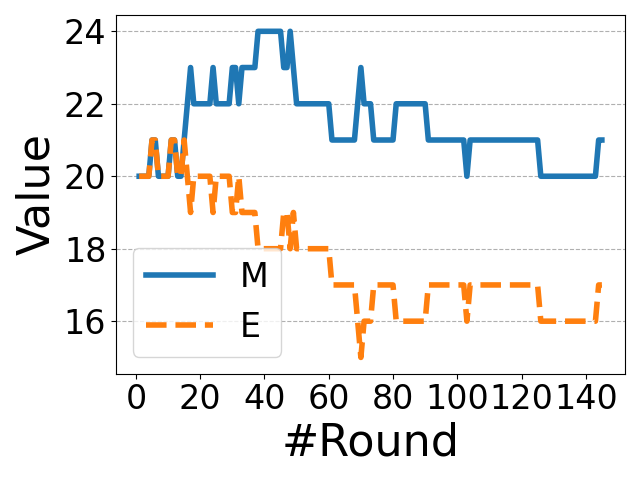}
}
\subfigure[(0.33-0.33-0-0.33)]{
    \includegraphics[width=0.18\textwidth]{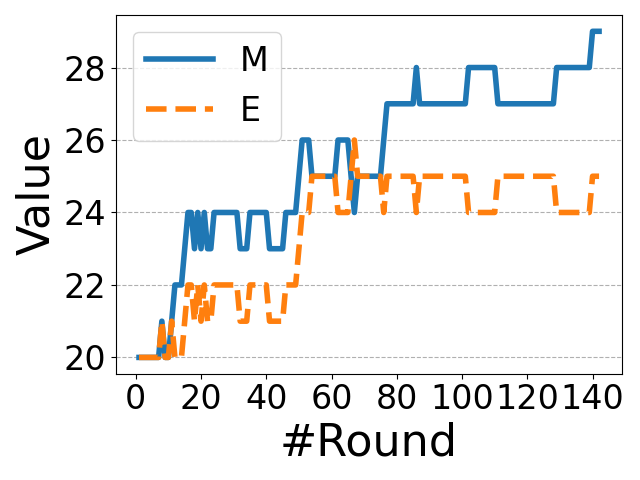}
}
\subfigure[(0.33-0-0.33-0.33)]{
    \includegraphics[width=0.18\textwidth]{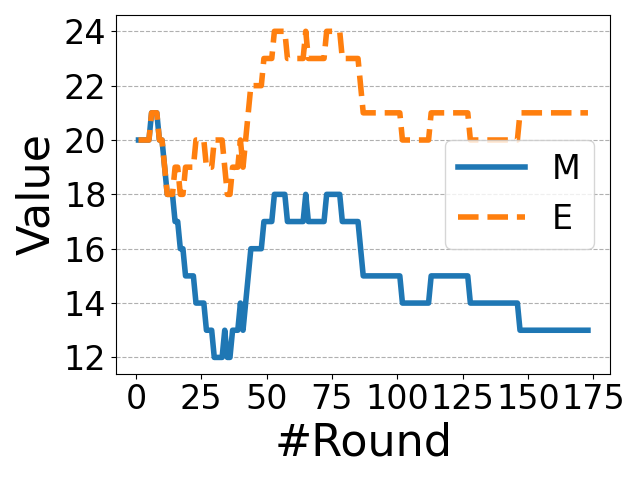}
}
\subfigure[(0-0.33-0.33-0.33)]{
    \includegraphics[width=0.18\textwidth]{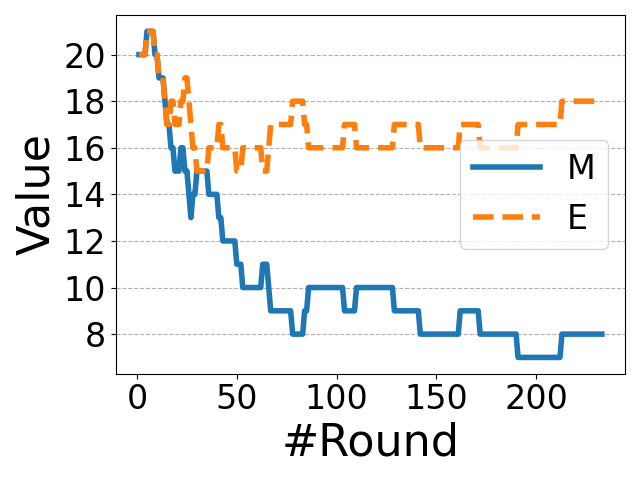}
}
\subfigure[(0.25-0.25-0.25-0.25)]{
    \includegraphics[width=0.18\textwidth]{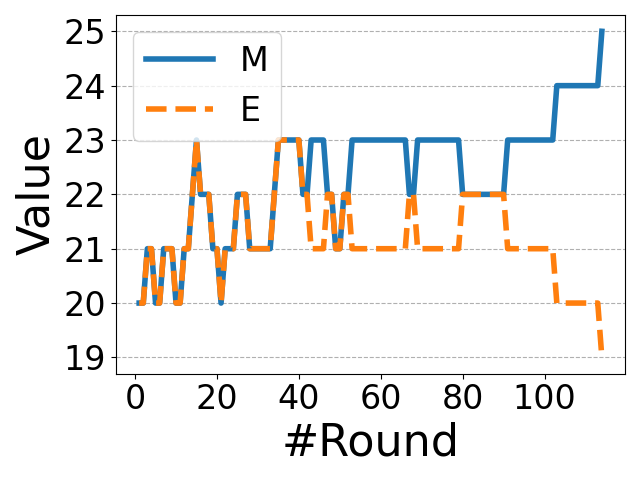}
}
}
    \caption{Illustrations of the number of participant M and the number of training passes E during FL training for different application preferences.}
    \label{fig:trace_illustration}
\end{figure*}

In this section, we provide the evaluation results of \shortname. First, we explain our experiment setup in Section \ref{sec:experiment_setup}. Next, we present the overall performance in Section \ref{sect:eva_overall}. Then, we conduct trace analysis in Section \ref{sec:trace_analysis}. Last, the penalty mechanism is studied in Section \ref{sec:penalty}.

\subsection{Experiment Setup}
\label{sec:experiment_setup}

\textbf{Benchmarks and Baseline}. 
We evaluate \shortname on three datasets: speech-to-command~\cite{speechToCommandDataset}, EMNIST~\cite{emnist}, and Cifar-100~\cite{cifar100}, and three aggregation methods: FedAvg~\cite{FedAvg17aistats}, FedNova~\cite{fedNova20neurips}, and FedAdagrad~\cite{fedyogi}. We set equal values for the combination of training preferences $\alpha$, $\beta$, $\gamma$ and $\delta$ (see the first column in Table~\ref{tab:eva_speech_to_command}). Therefore, for each dataset, we conduct 15 combinations of training preferences. We set target model accuracy for each dataset and measure CompT, TransT, CompL, and TransL for reaching the target model accuracy. We regard the practice of using fixed $M$ and $E$ as the baseline and compare \shortname to the baseline by calculating Eq.~(\ref{equ:compare}). In the evaluation, the positive performance means \shortname reduces the system overheads and the  negative performance means the degradation.
We implemented \shortname in PyTorch. All the experiments are conducted in a server with 24-GB Nvidia RTX A5000 GPUs.


\vspace{1mm}
\noindent
\textbf{Training Setup}. (1) \textit{speech-to-command} dataset. It classifies audio clips to 35 commands (e.g., `yes', `off'). We transform audio clips to 64-by-64 spectrograms and then downsize them to 32-by-32 gray-scale images. As officially suggested~\cite{speechToCommandDataset}, we use 2112 clients' data for training and the remaining 506 clients' data for testing. We set the mini-batch size to 5, considering that many clients have few data points. We use ResNet-10 and the target model accuracy of 0.8. (2) \textit{EMNIST} dataset. It classifies handwriting (28-by-28 gray-scale images) into 62 digits and letters (lowercase and uppercase). We split the dataset based on the writer ID. We randomly select 70\% writers' data for training and the remaining for testing. We use a Multiplayer Perception (MLP) model with one hidden layer~(200 neurons with ReLu activation). We set the mini-batch size to 10 and the target model accuracy of 0.7. (3) \textit{Cifar-100} dataset. It classifies 32-by-32 RGB images to 100 classes. We randomly split the dataset into 1200 users, where each user has 50 data points. Then, we randomly select 1000 users for training and the remaining 200 users for testing. We set the mini-batch size to 10. ResNet-18 is used, and the target model accuracy is set to 0.2 (due to our limited computational capability, we set a low threshold for Cifar-100).

\begin{table}[!t]
    \centering
\scalebox{0.9}{
    \centerline{
    \begin{tabular}{c |  c c  c}
\toprule
Dataset & Speech-command & EMNIST & Cifar-100 \\
Data Feature & Voice & Handwriting & Image \\ 
ML Model & ResNet-10 & 2-layer MLP & ResNet-10 \\[3pt]\hline 
\\[-5pt]
Performance & +22.48\% (17.97\%) & +8.48\% (5.51\%) & +9.33\% (5.47\%) \\
 \bottomrule
    \end{tabular}
    }
    }
\vspace{0.05in}
    \caption{Performance of \shortname for diverse datasets when FedAvg aggregation method is applied. }
    \label{tab:all_dataset}
\end{table}

For all datasets, we normalize the input images with the mean and the standard deviation of the training data before feeding them to models for training and testing. Both $M$ and $E$ are initially set to 20. \shortname is activated when the model accuracy is increased by at least 0.01 (i.e., $\epsilon = 0.01$). The penalty factor $D$ is set to 10. All results are averaged by three experiments.

\subsection{Overall Performance}
\label{sect:eva_overall}

\begin{table}[!t]
    \centering
\scalebox{0.9}{
    \centerline{
    \begin{tabular}{c | c c c}
    \toprule
Aggregator & FedAvg & FedNova & FedAdagrad \\
Performance & +22.48\% (17.97\%) & +23.53\% (6.64\%) & +26.75\% (6.10\%) \\
\bottomrule
    \end{tabular}
    }
    }
    \vspace{0.05in}
    \caption{Performance of \shortname for diverse aggregation algorithms. Speech-to-command dataset and ResNet-10 are used in this experiment.}
    \label{tab:all_aggregator}
\end{table}



\textbf{Results for Diverse Datasets}. 
Table \ref{tab:all_dataset} shows the overall performance of \shortname for different datasets when FedAvg is applied. We set the learning rate to 0.01 for the speech-to-command dataset and the EMNIST dataset, and 0.1 for the Cifar-100 dataset, all with the momentum of 0.9. We show the standard deviation in parenthesis. As shown, \shortname consistently improves the system performance across all the three datasets. In particular, \shortname reduces 22.48\% system overhead of the speech-to-command dataset compared to the baseline by averaging the 15 combinations of training preferences. 
We also observe that the FL training benefits more from \shortname if the training process needs more training rounds to converge. Our experiments with EMNIST~(small model) and Cifar100 (low target accuracy) only require a few dozens of training rounds to reach their target model accuracy, and thus their performance gains from \shortname are not significant. The observation is consistent with the decision-making process in \shortname, which increases/decreases hyper-parameters by only one at each step. We leave it as future work to augment \shortname to change hyper-parameters with adaptive degrees.  

\vspace{1mm}
\noindent
\textbf{Results for Different Aggregation Methods}. Table \ref{tab:all_aggregator} shows the overall performance of \shortname for different aggregation methods when we use the speech-to-command dataset and the ResNet-10 model. We set the learning rate to 0.1, $\beta_1$ to 0, and $\tau$ to 1e-3 in FedAdagrad. As shown, \shortname achieves consistent performance gain for diverse aggregation methods. In particular, \shortname reduces the system overhead of FedAdagrad by 26.75\%.  



\subsection{Trace Analysis}
\label{sec:trace_analysis}

We present the details of traces when the speech-to-command dataset and the FedAdagrad aggregation method are used. 
Table \ref{tab:eva_speech_to_command} tabulates the results, where we show the application preference ($\alpha$, $\beta$, $\gamma$, and $\delta$), the system overheads (CompT, TransT, CompL, and TransL), the final $M$ and $E$ when the training is finished, and the overall performance.
We report the average performance, as well as their standard deviations in parentheses. The first row is the baseline, which does not change hyper-parameters during the FL training. 
As we can see from Table \ref{tab:eva_speech_to_command}, \shortname can adapt to different training preferences. Specifically, \shortname reduces the system overhead up to 70.51\% when the application only cares about computation load (i.e., $\gamma = 1$).
Only one preference (0.33, 0.33, 0, 0.33) results in a slightly degraded performance. On average, \shortname improves the overall performance by 26.75\% for the speech-to-command dataset and the FedAdagrad aggregation method.

Fig.~\ref{fig:trace_illustration} illustrates one trace of $M$ and $E$ during the FL training for each application preference. The experiment is conducted with the speech-to-command dataset and the FedAdagrad aggregation method.
The experiment result clearly shows that \shortname can automatically adjust FL hyper-parameters during the FL training while respecting the application's preference on system overhead. We also observe that \shortname does not keep increasing or decreasing $M$ and $E$. Instead, in many cases, it intelligently tunes the $M$ and $E$ during the different phases of the FL training.



\subsection{Study of Penalty Mechanism}
\label{sec:penalty}

\begin{figure}[!t]
    \centering
    \includegraphics[width=3.2in]{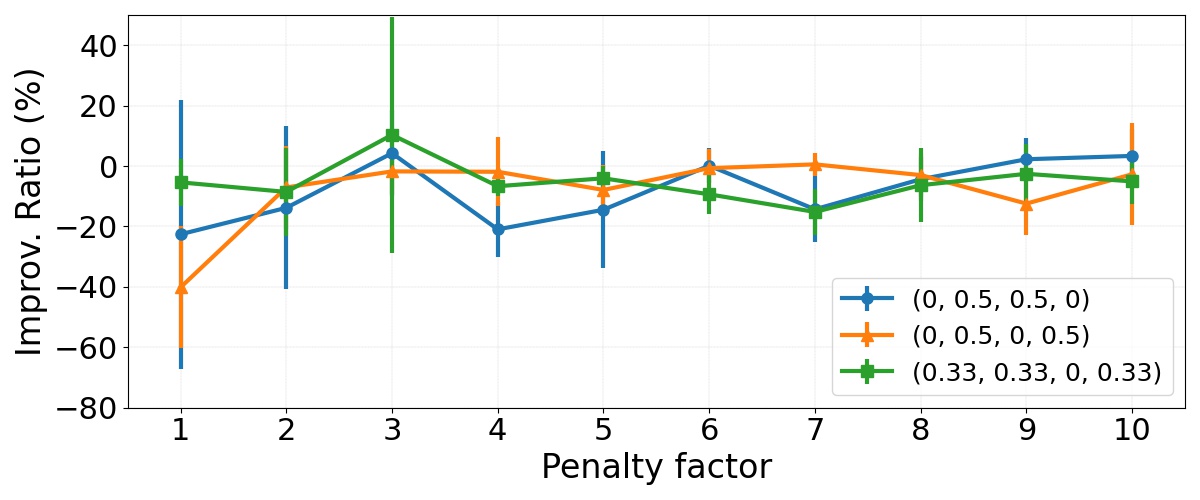}
    \caption{Performance of the degraded cases versus the penalty factor. Speech-to-command dataset and FedAvg is used in this experiment.}
    \label{fig:penalty_factor}
\end{figure}

We investigate our penalty mechanism by conducting experiments with the Google speech-to-command dataset and the FedAvg aggregation method. Without the penalty mechanism (equivalently $D=1$ since the penalty factor is a multiplier), we find that \shortname results in three degraded cases, i.e., the preferences of (0, 0.5, 0.5, 0), (0, 0.5, 0, 0.5), and (0.33, 0.33, 0, 0.33). 

First, we conduct experiments to explore whether the penalty mechanism can mitigate the degradation. Results are averaged over three runs. 
Fig.~\ref{fig:penalty_factor} plots the performance of \shortname for the degraded cases versus penalty factors,  where error bars represent the standard deviation. Although the penalty mechanism does not guarantee positive performance for the degraded cases, it mitigates the degradation. Fig.~\ref{fig:penalty_factor} also shows that \shortname remains stable for a moderate value of penalty factors. Empirically, we set the penalty factor to 10 in \shortname. 

\begin{figure}[!t]
    \centering
    \includegraphics[width=3.4in]{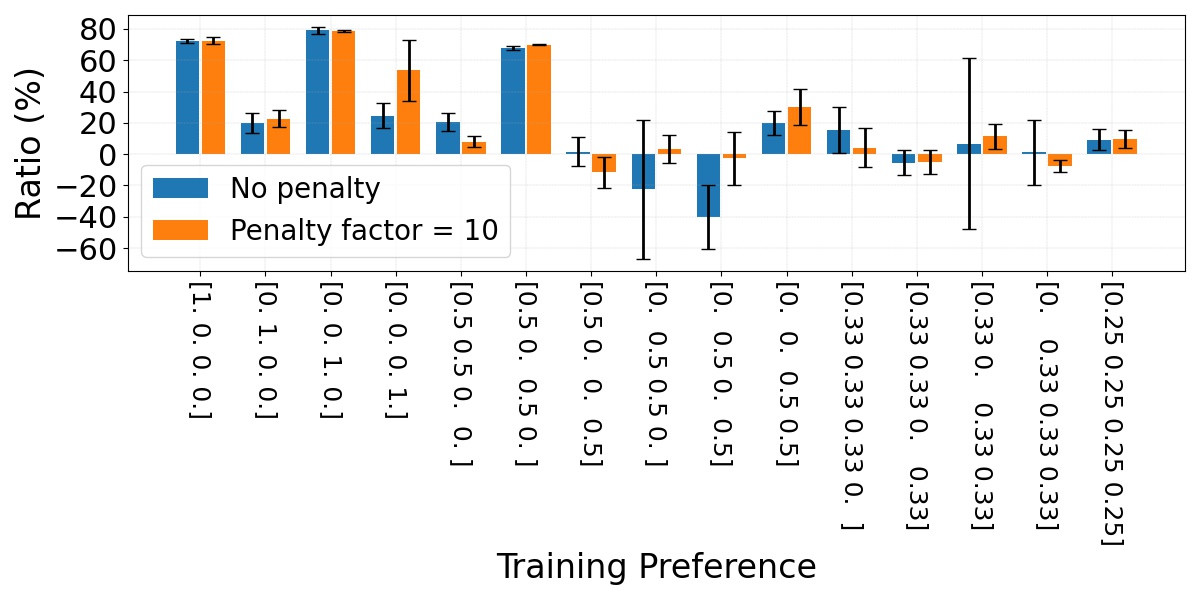}
    \caption{Performance comparison of \shortname without penalty mechanism and with penalty mechanism for the Speech-to-command dataset and FedAvg aggregation method.}
    \label{fig:overall_performance}
\end{figure}

Fig.~\ref{fig:overall_performance} compares the performance of \shortname when the penalty factor is 10 (full-fledged) versus 1 (no penalty) for different training preferences. Results are averaged by three runs and the error bars represent the standard deviation. Overall, \shortname increases the performance gain of the FedAvg aggregation method from 17.97\% to 22.48\%. In addition, \shortname with the penalty mitigates the highly degraded cases and thus is more practical than the non-penalty counterpart. We also observe that \shortname with penalty mechanism is more stable, with the averaged standard deviation of 7.77\% versus 14.14\% in the non-penalty counterpart.

\section{Discussion}
\label{sec:discuss}

\shortname has promising performance in tuning FL hyper-parameters. 
As one of the first work of its kind, \shortname has some limitations/opportunities that deserve further exploration.

\textit{Heterogeneous Devices}. This paper assumes that clients are
homogeneous, i.e., the same hardware and network. In practice, however, client devices are heterogeneous.
Measurements of the computation capabilities of mobile devices and their network throughput exhibit order-of-magnitude difference~\cite{ai_benchmark, mobiperf, fedscale}. 
As a result, for clients even with the same amount of local data points, they result in different computation and transmission costs. Currently, \shortname only tunes the system-wide hyper-parameters, which might not be optimal for heterogeneous devices that may require device-level hyper-parameter tuning. However, compared to the overwhelming and laboring effort of manually determining the FL hyper-parameters for each device, tuning system-wide hyper-parameter seems to be a more practical approach for FL practitioners.  
We leave it as future work to evaluate \shortname on heterogeneous devices.

\textit{Extensions}.  
Many FL algorithms have been proposed to tackle the limitations of FedAvg. \shortname could be extended to the following scenarios. (1) Participant selection. Compared to the random selection of participants, guided participant selection that considers clients' data utility and device utility can improve overall training performance~\cite{Oort21OSDI}. Popular alternatives are to only wait for participants that are finished before a deadline~\cite{tensorflowFL} or only wait for the first $M$ participants~\cite{fedAvgConverge20iclr}. (2) Adaptive training passes across participants. Due to the heterogeneity of clients, setting the same number of training passes $E$ for all participants in each training round is not optimal. To support different $E$ across participants, FedNova~\cite{fedNova20neurips} relies on re-weighting of aggregation while FedProx~\cite{fedprox} adds a proximal term to stabilize the convergence. We plan to incorporate these functionality to further improve the performance of \shortname. (3) Investigate more FL hyper-parameters. Although we only take the number of participants, the number of training passes, and the model complexity as examples to illustrate the workflow of \shortname, we believe \shortname is a general framework, which can be applied to other FL hyper-parameters, e.g., the mini-batch size.

\section{Conclusion}
\label{sec:conclusion}


FL involves high system overhead in the training process, which hinders its research and real-world deployment. We argue that optimizing system overhead for FL applications is extremely valuable. To this end, we propose \shortname to adjust FL hyper-parameters, catering to the application's training preferences automatically. 
Our evaluation results show that \shortname is general, lightweight,  flexible, and is able to significantly reduce system overhead in FL training. 


\ifCLASSOPTIONcaptionsoff
  \newpage
\fi

\bibliographystyle{unsrt}
\bibliography{reference}

%

\begin{IEEEbiography}[{\includegraphics[width=1in,height=1.25in,clip,keepaspectratio]{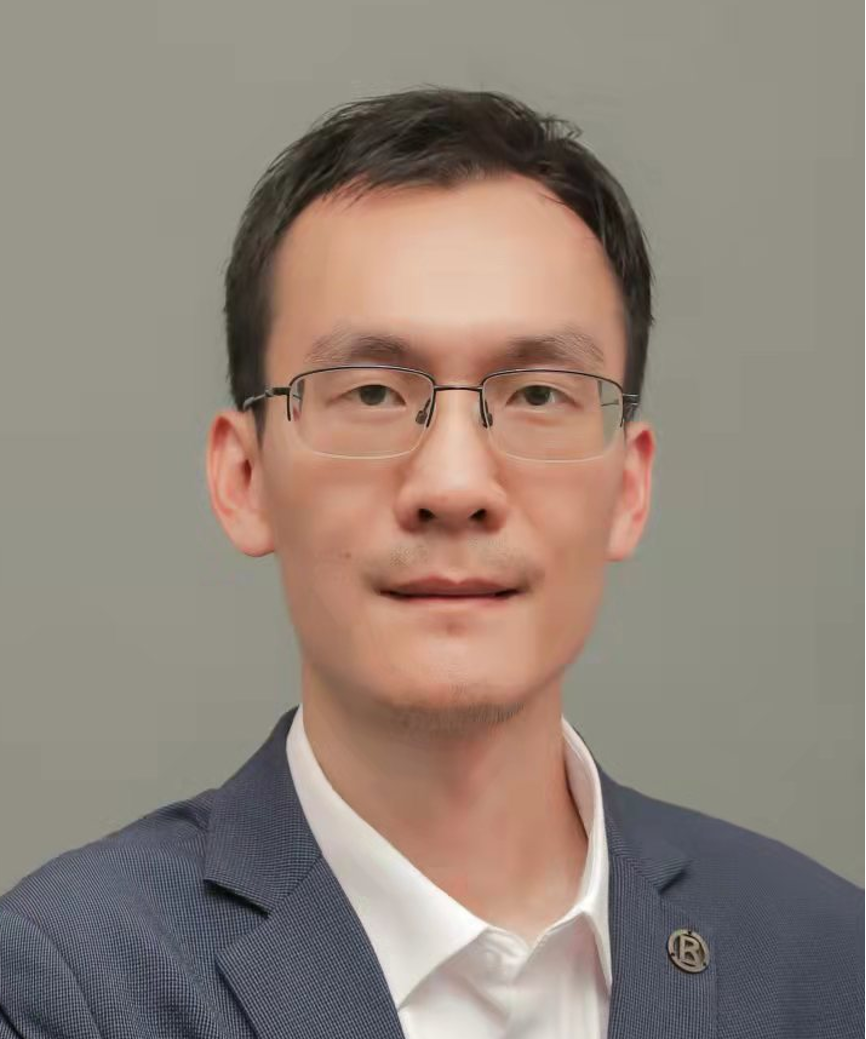}}]{Huanle Zhang}
is an associate professor in the School of Computer Science and Technology at Shandong University, China. He received his Ph.D. degree in computer science from the University of California, Davis (UC Davis), in 2020. He was employed as a postdoc at UC Davis from 2020 to 2022 and a project officer at Nanyang Technological University from 2014 to 2016. His research interests include data-centric AI, IoT, and mobile systems. 
\end{IEEEbiography}

\begin{IEEEbiography}[\raisebox{0.15in}{\includegraphics[width=1.1in,clip,keepaspectratio]{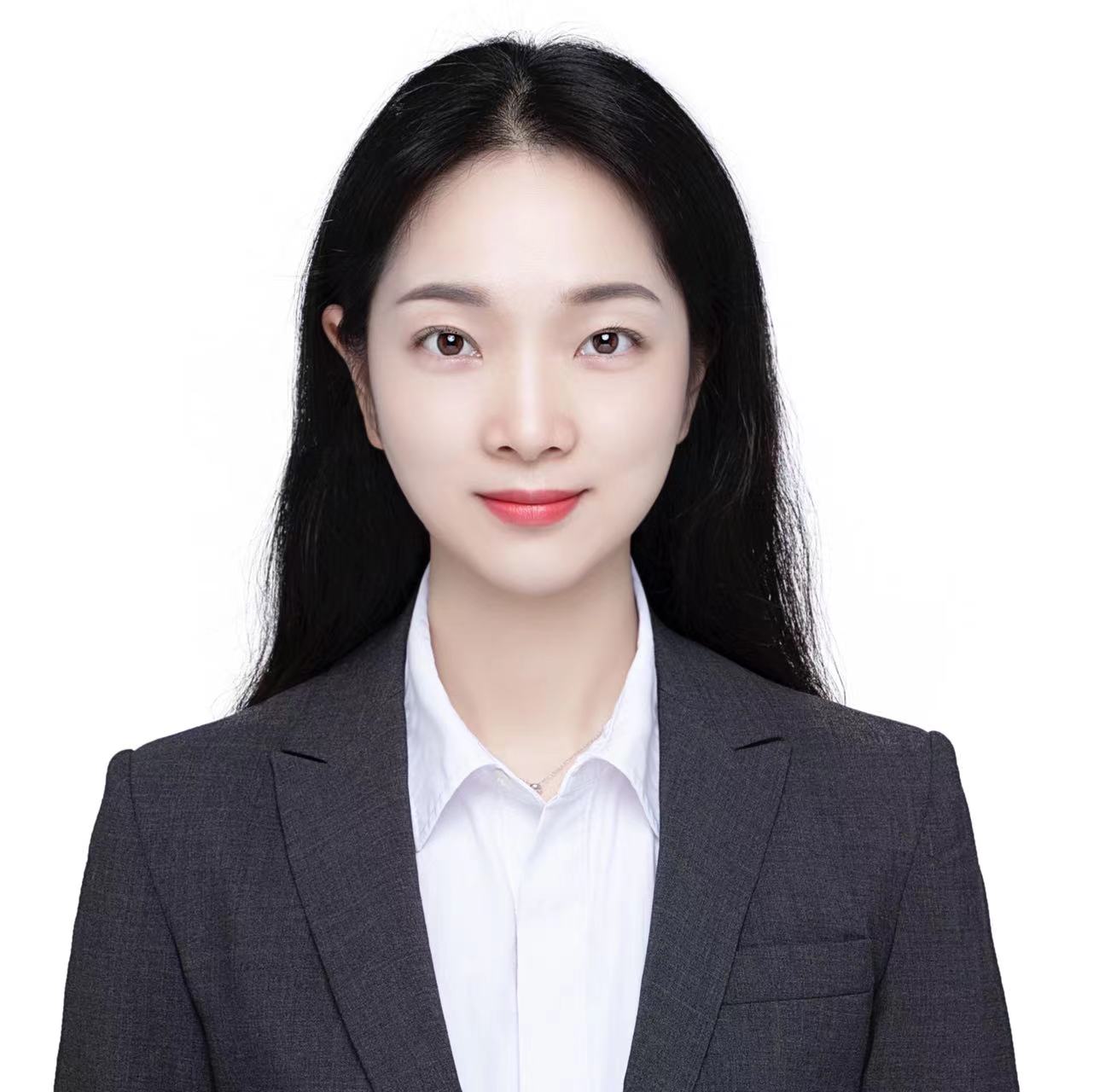}}]{Lei Fu} received her Ph.D. degree in Financial Information Engineering, Shanghai University of Finance and Economics in 2021. She is a postdoc in Fudan University and a quantitative investment analyst in Bank of Jiangsu, China. She received the Excellent Postdoc Award of Jiangsu in 2022. Her research interests include quantitative investment, 
macroeconomic computing, and financial technology.
\end{IEEEbiography}

\begin{IEEEbiography}[{\includegraphics[width=0.95in,clip,keepaspectratio]{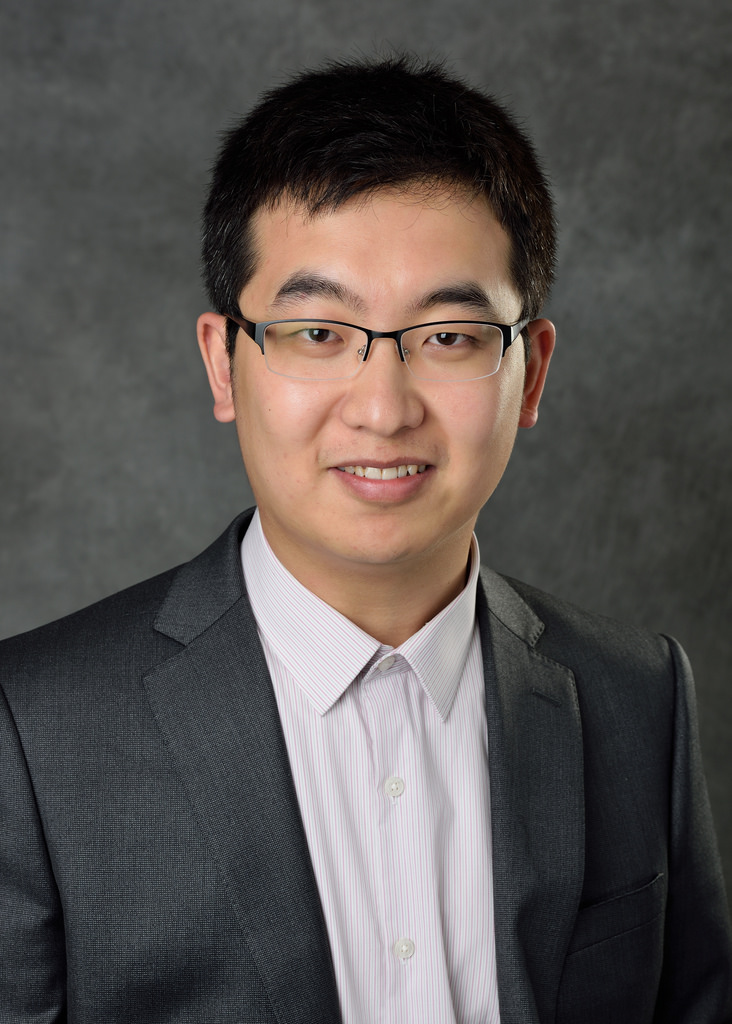}}]{Mi Zhang} is an associate professor in the Department of Computer Science and Engineering at The Ohio State University. He received his Ph.D. in Computer Engineering from the University of Southern California (USC). His research lies at the intersection of mobile/edge/IoT systems and machine learning. He is the recipient of seven best paper awards and nominations. He is also the recipient of the National Science Foundation CRII Award, Facebook Faculty Research Award, Amazon Machine Learning Research Award, and MSU Innovation of the Year Award. 
\end{IEEEbiography}

\begin{IEEEbiography}[{\includegraphics[width=1in,height=1.25in,clip,keepaspectratio]{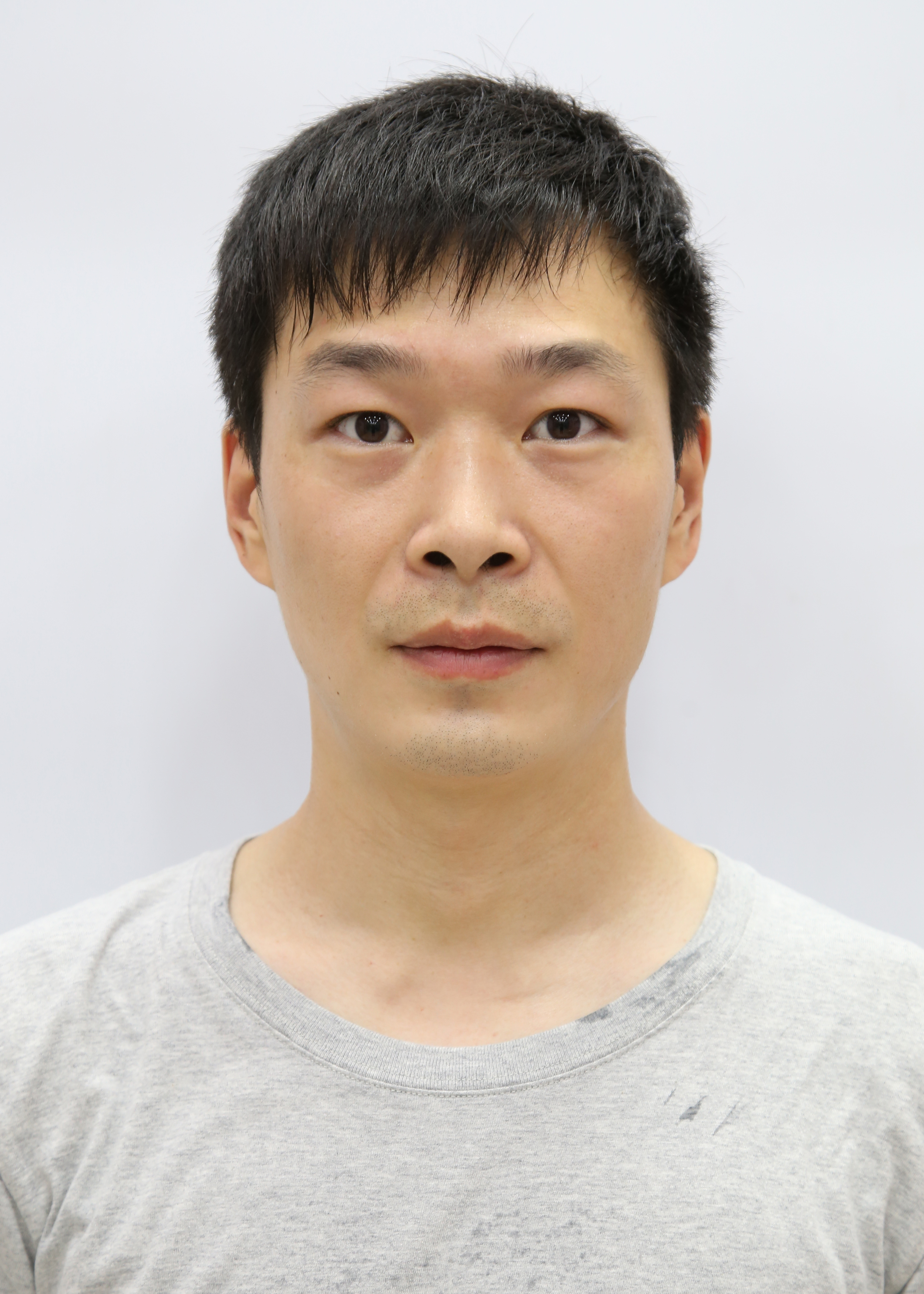}}]{Pengfei Hu} is a professor in School of Computer Science and Technology at Shandong University. He received Ph.D. in Computer Science from UC Davis. His Ph.D. supervisor is Prof. Prasant Mohapatra. His research interests are in the areas of cyber security, AI security, and mobile computing. He has published over 30 papers on these topics in premier conferences and journals. He served as TPC for numerous prestigious conferences.
\end{IEEEbiography}

\begin{IEEEbiography}[{\includegraphics[width=1in,height=1.25in,clip,keepaspectratio]{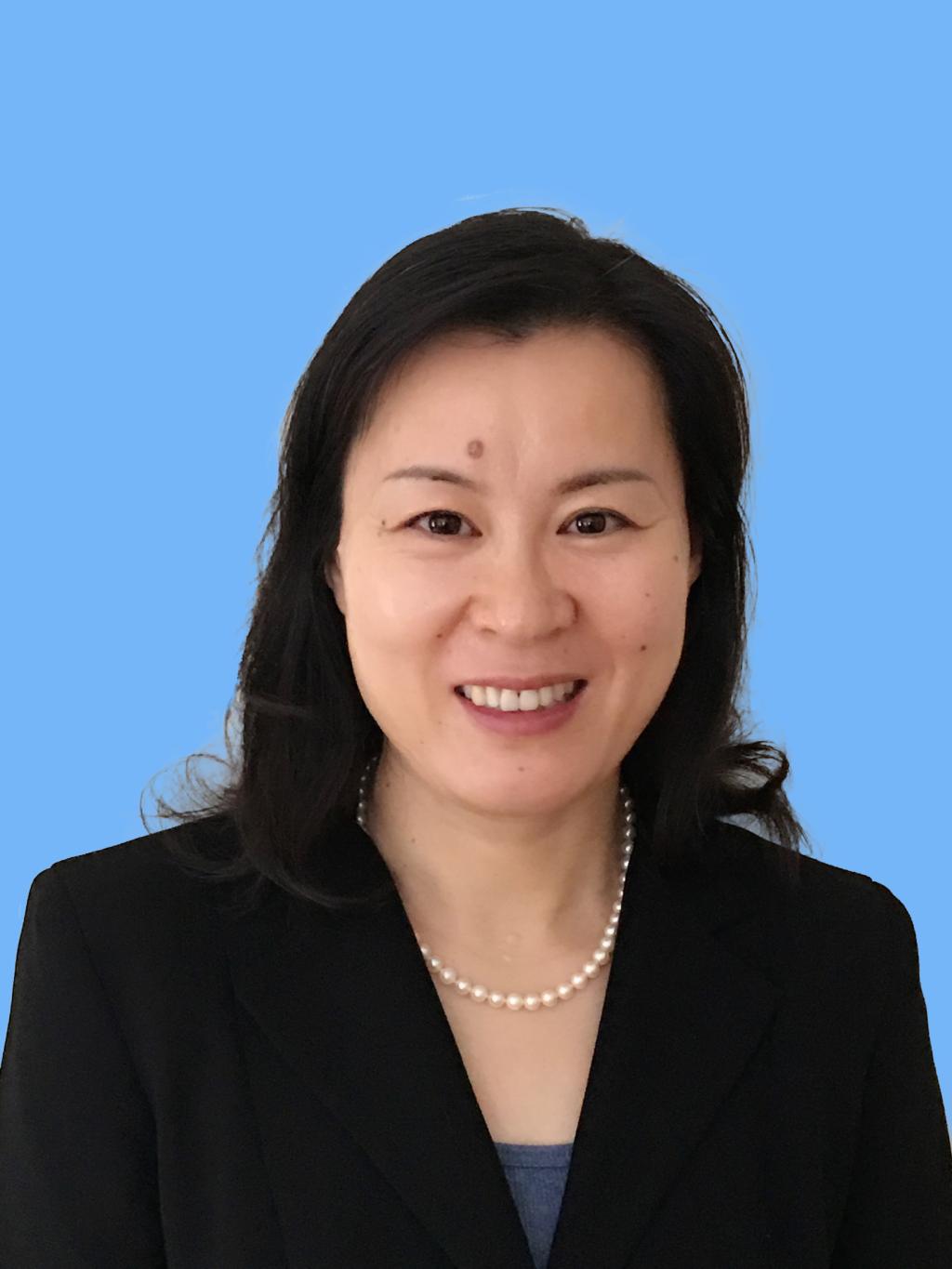}}]{Xiuzhen Cheng} received her M.S. and Ph.D. degrees in computer science from the University of Minnesota – Twin Cities in 2000 and 2002, respectively. She is a professor in the School of Computer Science and Technology, Shandong University. She worked as a program director for the US National Science Foundation (NSF) from April to October in 2006 (full time), and from April 2008 to May 2010 (part time). She received the NSF CAREER Award in 2004. She is a fellow of IEEE and a member of ACM.
\end{IEEEbiography}

\begin{IEEEbiography}[{\includegraphics[width=1.05in,clip,keepaspectratio]{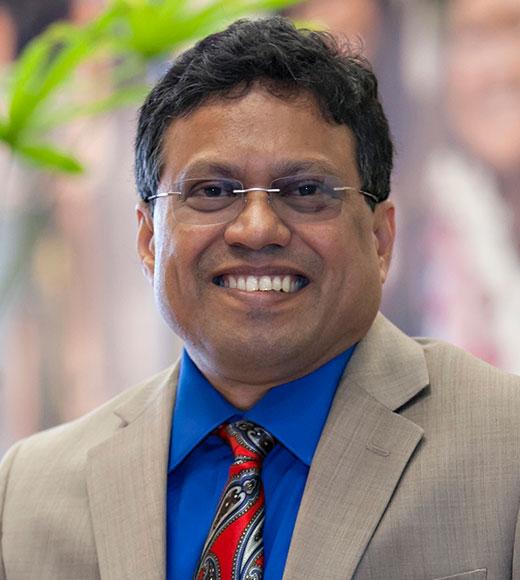}}]{Prasant Mohapatra} is serving as the Vice Chancellor for Research at University of California, Davis. He is also a Professor in the Department of Computer Science and served as the Dean and Vice-Provost of Graduate Studies at University of California, Davis during 2016-18. He was the editor-in-chief of the IEEE Transactions on Mobile
Computing. He has served on the editorial boards of the IEEE Transactions on Computers, the
IEEE Transactions on Mobile Computing, the IEEE Transaction on Parallel and Distributed
Systems, the ACM Journal on Wireless Networks, and Ad Hoc Networks. He is a fellow of
the IEEE and a fellow of the AAAS
\end{IEEEbiography}

\begin{IEEEbiography}[{\includegraphics[width=1.05in,clip,keepaspectratio]{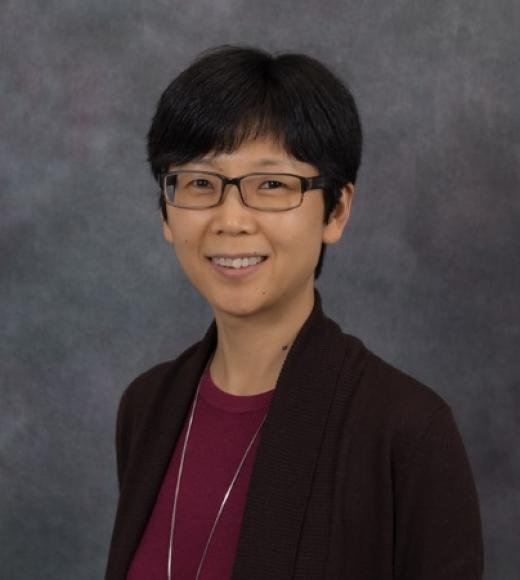}}]{Xin Liu} received her Ph.D. degree in electrical engineering from Purdue University in 2002. She is currently a Professor in Computer Science at the University of California, Davis. She received the Computer Networks Journal Best Paper of Year award in 2003 and NSF CAREER award in 2005. She became a Chancellor's Fellow in 2011. She is a co-PI and AI cluster co-lead for the USD 20M AI Institute for Next Generation Food Systems. She is a fellow of the IEEE.
\end{IEEEbiography}

\end{document}